\PassOptionsToPackage{unicode}{hyperref}
\PassOptionsToPackage{hyphens}{url}
\PassOptionsToPackage{dvipsnames,svgnames,x11names}{xcolor}
\documentclass[
  a4paper,
]{article}

\usepackage{amsmath,amssymb}
\usepackage{iftex}
\ifPDFTeX
  \usepackage[T1]{fontenc}
  \usepackage[utf8]{inputenc}
  \usepackage{textcomp} 
\else 
  \usepackage{unicode-math}
  \defaultfontfeatures{Scale=MatchLowercase}
  \defaultfontfeatures[\rmfamily]{Ligatures=TeX,Scale=1}
\fi
\usepackage{lmodern}
\ifPDFTeX\else  
\fi
\IfFileExists{upquote.sty}{\usepackage{upquote}}{}
\IfFileExists{microtype.sty}{
  \usepackage[]{microtype}
  \UseMicrotypeSet[protrusion]{basicmath} 
}{}
\makeatletter
\@ifundefined{KOMAClassName}{
  \IfFileExists{parskip.sty}{%
    \usepackage{parskip}
  }{
    \setlength{\parindent}{0pt}
    \setlength{\parskip}{6pt plus 2pt minus 1pt}}
}{
  \KOMAoptions{parskip=half}}
\makeatother
\usepackage{xcolor}
\usepackage[lmargin=19mm,rmargin=19mm,tmargin=19mm,bmargin=25mm]{geometry}
\ifx\paragraph\undefined\else
  \let\oldparagraph\paragraph
  \renewcommand{\paragraph}[1]{\oldparagraph{#1}\mbox{}}
\fi
\ifx\subparagraph\undefined\else
  \let\oldsubparagraph\subparagraph
  \renewcommand{\subparagraph}[1]{\oldsubparagraph{#1}\mbox{}}
\fi

\providecommand{\tightlist}{%
  \setlength{\itemsep}{0pt}\setlength{\parskip}{0pt}}\usepackage{longtable,booktabs,array}
\usepackage{calc} 
\usepackage{etoolbox}
\makeatletter
\patchcmd\longtable{\par}{\if@noskipsec\mbox{}\fi\par}{}{}
\makeatother
\IfFileExists{footnotehyper.sty}{\usepackage{footnotehyper}}{\usepackage{footnote}}
\makesavenoteenv{longtable}
\usepackage{graphicx}
\makeatletter
\def\maxwidth{\ifdim\Gin@nat@width>\linewidth\linewidth\else\Gin@nat@width\fi}
\def\maxheight{\ifdim\Gin@nat@height>\textheight\textheight\else\Gin@nat@height\fi}
\makeatother
\setkeys{Gin}{width=\maxwidth,height=\maxheight,keepaspectratio}
\makeatletter
\def\fps@figure{htbp}
\makeatother
\newlength{\cslhangindent}
\newlength{\csllabelwidth}
\newlength{\cslentryspacingunit} 
\newenvironment{CSLReferences}[2] 
 {
  \setlength{\parindent}{0pt}
  \ifodd #1
  \let\oldpar\par
  \def\par{\hangindent=\cslhangindent\oldpar}
  \fi
  \setlength{\parskip}{#2\cslentryspacingunit}
 }%
 {}
\usepackage{calc}

\makeatletter
\@ifpackageloaded{caption}{}{\usepackage{caption}}
\AtBeginDocument{%
\ifdefined\contentsname
  \renewcommand*\contentsname{Table of contents}
\else
  \newcommand\contentsname{Table of contents}
\fi
\ifdefined\listfigurename
  \renewcommand*\listfigurename{List of Figures}
\else
  \newcommand\listfigurename{List of Figures}
\fi
\ifdefined\listtablename
  \renewcommand*\listtablename{List of Tables}
\else
  \newcommand\listtablename{List of Tables}
\fi
\ifdefined\figurename
  \renewcommand*\figurename{Figure}
\else
  \newcommand\figurename{Figure}
\fi
\ifdefined\tablename
  \renewcommand*\tablename{Table}
\else
  \newcommand\tablename{Table}
\fi
}
\@ifpackageloaded{float}{}{\usepackage{float}}
\floatstyle{ruled}
\@ifundefined{c@chapter}{\newfloat{codelisting}{h}{lop}}{\newfloat{codelisting}{h}{lop}[chapter]}
\floatname{codelisting}{Listing}

\makeatother
\makeatletter
\@ifpackageloaded{caption}{}{\usepackage{caption}}
\@ifpackageloaded{subcaption}{}{\usepackage{subcaption}}
\makeatother
\makeatletter
\@ifpackageloaded{tcolorbox}{}{\usepackage[skins,breakable]{tcolorbox}}
\makeatother
\makeatletter
\@ifundefined{shadecolor}{\definecolor{shadecolor}{rgb}{.97, .97, .97}}
\makeatother
\ifLuaTeX
  \usepackage{selnolig}  
\fi
\IfFileExists{bookmark.sty}{\usepackage{bookmark}}{\usepackage{hyperref}}
\IfFileExists{xurl.sty}{\usepackage{xurl}}{} 
\hypersetup{
  pdftitle={Statistical inference using machine learning and classical techniques based on accumulated local effects (ALE)},
  pdfauthor={Chitu Okoli \textbar{} SKEMA Business School -- Université Côte d'Azur, Paris},
  pdfkeywords={accumulated local effects (ALE), machine
learning, interpretable machine learning (IML), explainable artificial
intelligence (XAI), statistical inference, confidence
intervals, bootstrap, effect size},
  colorlinks=true,
  linkcolor={blue},
  filecolor={Maroon},
  citecolor={Blue},
  urlcolor={Blue},
  pdfcreator={LaTeX via pandoc}}

\title{Statistical inference using machine learning and classical
techniques based on accumulated local effects (ALE)}
\author{Chitu Okoli \textbar{} SKEMA Business School -- Université Côte
d'Azur, Paris}
\date{2024-02-13}

\begin{document}
\maketitle
\begin{abstract}
Accumulated Local Effects (ALE) is a model-agnostic approach for global
explanations of the results of black-box machine learning (ML)
algorithms. There are at least three challenges with conducting
statistical inference based on ALE: ensuring the reliability of ALE
analyses, especially in the context of small datasets; intuitively
characterizing a variable's overall effect in ML; and making robust
inferences from ML data analysis. In response, we introduce innovative
tools and techniques for statistical inference using ALE, establishing
bootstrapped confidence intervals tailored to dataset size and
introducing ALE effect size measures that intuitively indicate effects
on both the outcome variable scale and a normalized scale. Furthermore,
we demonstrate how to use these tools to draw reliable statistical
inferences, reflecting the flexible patterns ALE adeptly highlights,
with implementations available in the `ale' package in R. This work
propels the discourse on ALE and its applicability in ML and statistical
analysis forward, offering practical solutions to prevailing challenges
in the field.
\end{abstract}
\ifdefined\Shaded\renewenvironment{Shaded}{\begin{tcolorbox}[breakable, sharp corners, boxrule=0pt, interior hidden, borderline west={3pt}{0pt}{shadecolor}, enhanced, frame hidden]}{\end{tcolorbox}}\fi

\hypertarget{introduction}{%
\section{Introduction}\label{introduction}}

Accumulated Local Effects (ALE) were initially developed as a
model-agnostic approach for global explanations of the results of
black-box machine learning (ML) algorithms (Apley and Zhu 2020). ALE has
a key advantage over other approaches like partial dependency plots
(PDP) and SHapley Additive exPlanations (SHAP): its values represent a
clean functional decomposition of the model (Molnar, Casalicchio and
Bischl 2020). As such, ALE values are not affected by the presence or
absence of interactions among variables in a mode. Moreover, its
computation is relatively rapid (Molnar 2022).

Despite the potential of ALE, at least three challenges in interpretable
machine learning (IML) remain unresolved, despite numerous incremental
attempts. Firstly, the reliability of results derived from ALE analyses
is brought into question, particularly when considering the prevalent
data-only bootstrapping approach (Flora 2023; Jumelle, Kuhn-Regnier and
Rajaratnam 2020), which, especially in the context of smaller datasets
that preclude a training-test split, potentially risks overfitting and
undermines the generalizability of the findings.

This important issue is primarily a concern for small datasets, so it is
pertinent to ask how small is ``small''? From the perspective of this
article, the key issue at stake is that applying the training-test split
that is common in ML is a crucial technique for increasing the
generalizability of data analysis. So, the question becomes focused on,
``How small is too small for a training-test split for ML analysis?'' We
could consider a general principle that ML requires at least 200 rows of
data for each predictor variable. So, for example, with five input
variables, we would need at least 1,000 rows of data. But this number
refers not to the size of the entire dataset but to the minimum size of
the training subset. So, with an 80-20 split on the full dataset (that
is, 80\% training set), we would need at least 1,000 rows for the
training set and another 250 rows for the test set, for a minimum of
1,250 rows. (And if we were to carry out hyperparameter tuning with
cross-validation on that training set, then we would need even more
data.) From these considerations, we suggest that most datasets of less
than 2,000 rows are probably ``small''. Indeed, even many datasets that
are more than 2,000 rows are nonetheless ``small''.

This consideration is pertinent because ALE is a valuable technique for
visually characterizing the relationships between predictors and
outcomes for any model, not just for large datasets typical in ML, but
also for smaller datasets typical in statistical analysis. It is often
inappropriate to transfer the ALE analysis techniques that assume large
datasets to smaller datasets that need specialized treatment.

A second ongoing challenge concerns how to characterize a variable's
overall effect. While effect sizes are extensively treated in
statistical analysis based on the general linear model, only recently
have ML researchers started trying to develop model-agnostic measures
that can characterize the results of any ML analysis. These initial
attempts, while promising, do not always show the effects of individual
variables (Molnar, Casalicchio and Bischl 2020) and, when they do, are
often not intuitively interpretable (Molnar, Casalicchio and Bischl
2020; Lötsch and Ultsch 2020; Friedman and Popescu 2008).

Third, even with reliable effect size measures, it is not clear how to
make reliable inferences from data analysis based on ML. Messner (2023)
has made an admirable initial attempt towards hypothesis testing based
on ML analysis, but his framework nonetheless masks some of the fine
nuances that ALE hopes to uncover with its visual portrayal of flexible
relationships.

In response to these challenges, the research objective of this article
is to introduce tools and techniques for statistical inference using
machine learning and classical statistical techniques based on ALE. We
address trusting the reliability of analysis results by creating
bootstrapped confidence intervals for ALE using techniques appropriate
to the size of the dataset. To characterize the overall effects of a
predictor variable on its outcome variable, we create a set of ALE
effect size measures that intuitively indicate the effect either on the
scale of the outcome variable or on a normalized scale that is
comparable across datasets. However, to avoid simplistic summaries of
overall effects that might hide important details, we demonstrate how to
use these tools to make reliable conclusions of statistical inference
that reflect the flexible patterns that ALE can so capably highlight.

We conduct all analyses using the R package \texttt{ale} (Okoli 2023),
which was specifically developed to extend ALE with the functionality we
describe in this article. Indeed, this article is largely an explanation
of the scientific background of the \texttt{ale} package.

The rest of this article is organized as follows. In the ``Related
work'' section, we delve into the existing literature and software
implementations of ALE, exploring aspects such as ALE confidence
intervals, bootstrapping, effect size measures in machine learning, and
inference from analysis results, while also identifying opportunities
for improvement in the current methodologies. Subsequently,
``Illustrative datasets and models'' introduces two distinct datasets
and corresponding models---a random forest model for diamond prices and
a generalized additive model for mathematics achievement scores---to
provide a practical context for the ensuing analyses. The section on
``Bootstrapping of accumulated local effects'' elucidates the
methodologies of data-only bootstrapping, particularly focusing on a
large dataset, and introduces model bootstrapping, with a spotlight on a
smaller dataset and its application to ALE. Moving forward, ``ALE effect
size measures'' introduces and explores novel effect size measures, both
on the scale of the \emph{y} outcome variable and in a normalized form,
while also discussing the implications and applications of random
variables. In ``Statistical inference with ALE,'' we navigate through
classical statistical inference, explore ALE data structures, and delve
into bootstrap-based inference with ALE, culminating in a discussion on
confidence regions and random variables. Finally, the ``Discussion''
section encapsulates the contributions and practical implications of the
methodologies and insights presented, providing a comprehensive overview
and concluding remarks on the potential trajectories and applications of
the enhanced ALE methodologies in machine learning model interpretation
and analysis.

\hypertarget{related-work}{%
\section{Related work}\label{related-work}}

In the seminal work of Apley and Zhu (2020), the concept of Accumulated
Local Effects (ALE) was introduced as a global explanation technique for
IML. ALE provides functionality akin to Partial Dependence Plots (PDP),
both of which graphically delineate the relationship between a single
input variable \emph{x} and the outcome variable \emph{y}, thereby
illustrating non-linear and flexible variant relationships. A noteworthy
enhancement of ALE over PDP is its resilience to the interactions
between variables. A secondary, yet non-negligible advantage of
utilizing the ALE algorithm is its reduced computational expense, which
serves as a substantial practical incentive for machine learning
scientists to incorporate ALE visualizations throughout their analytical
processes.

There have been a few recent extensions to fine-tune the original
algorithm. Gkolemis et al. (2023) present Robust and Heterogeneity-Aware
Accumulated Local Effects (RHALE), a technique that endeavours to
address a notable limitation inherent in the original ALE algorithm's
approach to binning numeric data according to quantiles. A pivotal
challenge resides in the absence of indicators for the heterogeneity of
data within each quartile. Consequently, the ALE plotted line may not
accurately represent the data. To mitigate this, RHALE considers the
standard deviation of data within each prospective bin when determining
the ALE bin boundaries. Gkolemis, Dalamagas and Diou (2023) present
Differential Accumulated Local Effects (DALE) to address two prominent
limitations of the original ALE formulation. Firstly, they argue that
the original ALE does not scale effectively to high-dimensional data.
Secondly, they note that with smaller samples, the original ALE
calculation may not be representative of out-of-distribution sampling.
DALE is an adjusted algorithm for calculating ALE that attempts to
enhance both the scalability and representational accuracy of the
technique

Although our present article focuses on the original ALE algorithm, its
findings can probably be readily extrapolated to such extensions.
Indeed, the kinds of extensions in which we are interested have not been
documented in scholarly articles that we could find but are rather found
to various extents in software packages that implement ALE. Thus, our
review in this section largely surveys software implementations in
addition to published literature.

\hypertarget{software-implementations-of-ale}{%
\subsection{Software implementations of
ALE}\label{software-implementations-of-ale}}

Since the initial \texttt{ALEPlot} reference package in R was released
in 2018 (Apley 2018), implementations in various programming languages
have translated or extended the initial concept. \texttt{ALEPython}
(Jumelle, Kuhn-Regnier and Rajaratnam 2020) and \texttt{PyALE} (Jomar
2023) in Python have been specifically dedicated to implementing ALE,
sometimes with extensions. Most, however, are more general IML packages
that include ALE among other techniques. These include \texttt{iml}
(Molnar and Schratz 2022) in R; \texttt{scikit-explain} (Flora 2023) and
\texttt{Alibi} (Seldon Technologies 2023) in Python; \texttt{DALEX}
(Biecek, Maksymiuk and Baniecki 2023) in R and Python; and the
\texttt{Interpretation} (RapidMiner 2023) extension for RapidMiner, a
Java-based no-code machine learning platform. In addition to these, this
present article introduces \texttt{ale} (Okoli 2023), an R package
dedicated to the implementation and extension of ALE. In particular,
\texttt{ale} aims to resolve some of the issues that we describe
subsequently in this section. In Table~\ref{tbl-ale-packages}, we list
some key characteristics of these packages, focusing on features that
are pertinent to the subject of this present article.

\hypertarget{tbl-ale-packages}{}
\begin{longtable}[]{@{}
  >{\raggedright\arraybackslash}p{(\columnwidth - 12\tabcolsep) * \real{0.0821}}
  >{\raggedright\arraybackslash}p{(\columnwidth - 12\tabcolsep) * \real{0.2718}}
  >{\raggedleft\arraybackslash}p{(\columnwidth - 12\tabcolsep) * \real{0.0872}}
  >{\raggedright\arraybackslash}p{(\columnwidth - 12\tabcolsep) * \real{0.0769}}
  >{\raggedright\arraybackslash}p{(\columnwidth - 12\tabcolsep) * \real{0.1179}}
  >{\raggedright\arraybackslash}p{(\columnwidth - 12\tabcolsep) * \real{0.1128}}
  >{\raggedright\arraybackslash}p{(\columnwidth - 12\tabcolsep) * \real{0.2256}}@{}}
\caption{\label{tbl-ale-packages}Software packages that implement
ALE}\tabularnewline
\toprule\noalign{}
\begin{minipage}[b]{\linewidth}\raggedright
Primary focus
\end{minipage} & \begin{minipage}[b]{\linewidth}\raggedright
Package
\end{minipage} & \begin{minipage}[b]{\linewidth}\raggedleft
Latest release
\end{minipage} & \begin{minipage}[b]{\linewidth}\raggedright
Language
\end{minipage} & \begin{minipage}[b]{\linewidth}\raggedright
Confidence intervals
\end{minipage} & \begin{minipage}[b]{\linewidth}\raggedright
Bootstrap type
\end{minipage} & \begin{minipage}[b]{\linewidth}\raggedright
ALE statistics
\end{minipage} \\
\midrule\noalign{}
\endfirsthead
\toprule\noalign{}
\begin{minipage}[b]{\linewidth}\raggedright
Primary focus
\end{minipage} & \begin{minipage}[b]{\linewidth}\raggedright
Package
\end{minipage} & \begin{minipage}[b]{\linewidth}\raggedleft
Latest release
\end{minipage} & \begin{minipage}[b]{\linewidth}\raggedright
Language
\end{minipage} & \begin{minipage}[b]{\linewidth}\raggedright
Confidence intervals
\end{minipage} & \begin{minipage}[b]{\linewidth}\raggedright
Bootstrap type
\end{minipage} & \begin{minipage}[b]{\linewidth}\raggedright
ALE statistics
\end{minipage} \\
\midrule\noalign{}
\endhead
\bottomrule\noalign{}
\endlastfoot
ALE & \textbf{ALEPlot} (Apley 2018) & 2018 & R & No & N/A & None \\
ALE & \textbf{ALEPython} (Jumelle, Kuhn-Regnier and Rajaratnam 2020) &
2020 & Python & Monte Carlo & data-only & None \\
IML & \textbf{iml} (Molnar and Schratz 2022) & 2022 & R & No & N/A &
None \\
IML & \textbf{DALEX} (Biecek, Maksymiuk and Baniecki 2023) & 2023 & R
and Python & No & N/A & None \\
ALE & \textbf{PyALE} (Jomar 2023) & 2023 & Python & T-statistic & N/A &
None \\
IML & \textbf{Interpretation} (RapidMiner 2023) & 2023 & RapidMiner & No
& N/A & None \\
IML & \textbf{Alibi} (Seldon Technologies 2023) & 2023 & Python & No &
N/A & None \\
IML & \textbf{scikit-explain} (Flora 2023) & 2023 & Python & Bootstrap &
data-only & \begin{minipage}[t]{\linewidth}\raggedright
\begin{itemize}
\item
  Friedman H-statistic for interactions
\item
  Interaction strength (IAS)
\item
  Main effect complexity (MEC)
\end{itemize}
\end{minipage} \\
ALE & \textbf{ale (introduced in this article)} (Okoli 2023) & 2023 & R
& Bootstrap & data-only and model &
\begin{minipage}[t]{\linewidth}\raggedright
\begin{itemize}
\item
  ALE deviation (ALED)
\item
  ALE ranger (ALER)
\item
  Normalized ALED (NALED)
\item
  Normalized ALER (NALER)
\end{itemize}
\end{minipage} \\
\end{longtable}

In the following subsections, we discuss publications and software
implementations that develop confidence intervals, bootstrapping, effect
size measures, and inference, almost always with ALE.

\hypertarget{ale-confidence-intervals-and-bootstrapping}{%
\subsection{ALE confidence intervals and
bootstrapping}\label{ale-confidence-intervals-and-bootstrapping}}

The initial \texttt{ALEPlot} implementation did not feature confidence
intervals, but several packages have recognized their importance and
have thus extended ALE with this feature, implemented in different ways.
The simplest approach is that employed by \texttt{PyALE}, which
constructs intervals based on the t statistic, using the standard
deviation of the ALE values to construct standard errors. However,
despite the other merits of that implementation, we cannot find any
basis to assume that ALE values are distributed according to a t
distribution, or by any other parametric distribution for that matter.
Only bootstrap-based confidence intervals are appropriate for data like
ALE whose distribution cannot be generalized to any predetermined
pattern (Tong, Saminathan and Chang 2017).

Although most readers are likely familiar with the bootstrap algorithm,
it is worth repeating it here to highlight the complexities that ALE
presents. The classic bootstrap approach involves creating multiple
samples of the original dataset. Each bootstrap sample draws rows at
random from the original dataset, as many new rows as there are rows in
the original dataset. Crucially, sampling is done with replacement so
that in each bootstrap sample, some of the original rows are repeated a
random number of times and some might not occur at all. The analyst
decides how many bootstrap samples they want. With the bootstrap
samples, the desired aggregate statistics are computed, such as the mean
or standard deviation across all bootstrap samples.

For ALE bootstrapping, the aggregation goal across the bootstrap samples
is to calculate the ALE values. The ALE calculation depends first on
establishing ALE bins or intervals for an \emph{x} predictor variable
and then an ALE \emph{y} estimate is calculated for each ALE \emph{x}
interval. What is particularly tricky with bootstrapping ALE
calculations is that the intervals are calculated directly from the
data, so each time the data is scrambled with a bootstrap sample, the
ALE \emph{x} intervals necessarily change. Even with large datasets,
because the intervals for numeric \emph{x} predictors are calculated
from the quantiles of the particular sets of values, each bootstrap
sample will produce distinct intervals, which cannot be combined in any
way across bootstrap samples. The problem persists even with categorical
\emph{x} predictors. Whereas with sufficiently large datasets each
bootstrap sample will likely include at least a few rows that represent
each category of the \emph{x} predictor, with small datasets, it is
often the case that at least some bootstrap samples might not include
every category from the original dataset. In such cases, ALE values
cannot be combined across bootstrap samples that do not share identical
intervals.

In their implementations of bootstrapped confidence intervals, the
\texttt{scikit-explain} (Flora 2023) and \texttt{ale} (Okoli 2023)
packages resolve this challenge by using the full dataset to establish
the ALE \emph{x} intervals and then they apply these fixed intervals in
calculating the ALE \emph{y} values for each bootstrap sample. Thus,
bootstrap averages and quantiles can be calculated across each common
interval across all bootstrap samples. In general, we call this standard
approach \textbf{``data-only bootstrapping''} because the dataset is
bootstrapped, but not the model itself. This approach might also be
called model-invariant bootstrapping or case resampling.

The \texttt{ALEPython} package (Jumelle, Kuhn-Regnier and Rajaratnam
2020) adopts a similar approach in their adapted bootstrap confidence
intervals that they call ``Monte Carlo''. This is essentially the same
as the standard bootstrap (sampling with replacement) except that
instead of constructing samples of the same size as the original
dataset, they only sample a fixed fraction of the dataset (10\% by
default, customizable by the user). Other than that detail, these three
packages essentially employ the same approach.

\hypertarget{effect-size-measures-for-machine-learning}{%
\subsection{Effect size measures for machine
learning}\label{effect-size-measures-for-machine-learning}}

Although effect size measures are widely used in statistical analyses,
usually with smaller datasets, they are not as widely used in machine
learning. However, there are a few notable examples. Friedman and
Popescu (2008) introduced a metric known as the H-statistic, designed to
quantify the intensity of interactions among variables in datasets. It
comes in two forms: a pairwise version that reveals the strength of
two-way interactions between feature pairs, and another that assesses
the interaction strength of a single feature with any other feature in
the dataset or model. Lötsch and Ultsch (2020) have formulated an effect
size metric, drawing parallels to Cohen's D used in statistical
evaluations of group differences. Their innovation, tailored for the
machine learning context, not only considers the central tendencies of
groups but also the variations within and across these groups, providing
a more nuanced measure.

Molnar, Casalicchio and Bischl (2020) delve into interpreting models
through a process they refer to as functional decomposition. This
approach involves dissecting a model's predictive elements into three
distinct components: a constant element, the main effects of individual
variables, and interactions between variables. For evaluating main
effects, their methodology hinges on ALE, leading to the formulation of
a metric termed Main Effect Complexity (MEC). In essence, MEC gauges the
degree to which a variable's relationship with the outcome diverges from
a linear trend. Regarding the interaction component, they employ a
metric known as Interaction Strength (IAS), which serves to signify the
robustness of a feature's interaction with others.

Regarding MEC, two distinct calculations exist. Initially, MEC is
determined for each feature individually, assessing the extent to which
the relationship between a feature and the outcome deviates from a
linear representation. Subsequently, an overall MEC is computed as the
mean of individual MECs for all features within the model, with this
aggregate value typically being the primary focus. Conversely, IAS is
consistently calculated for the entire set of features in the model. It
essentially measures what remains unaccounted for when the average main
effects of all features are considered. Therefore, IAS represents a
collective metric for the entire model, rather than a measure of
interaction for individual features. The \texttt{scikit-explain} package
(Flora 2023) provides Friedman's H-statistic, MEC, and IAS effect size
measures.

Messner (2023) adapts Cohen's \(f^2\) from classical statistical
analysis to quantify effect strength in ML contexts. He diverges from
the conventional version based on \(R^2\) used in ordinary least squares
regression to an adapted version he calls \(f^2_v\), based on mean
squared error (MSE). \(f^2_v\), a unit-free measure that ranges from 0
with no upper bound, can be compared across datasets. Additionally,
Messner evaluates the monotonicity of an effect---whether an effect
consistently moves in one direction---and its directionality, utilizing
p-values derived from the Mann-Kendall test and Theil-Sen slope,
respectively. These tests determine if the effect is unidirectional and
if it is increasing or decreasing, with the magnitude of change assessed
via \(f^2_v\).

\hypertarget{inference-from-analysis-results}{%
\subsection{Inference from analysis
results}\label{inference-from-analysis-results}}

One of the biggest differences between supervised ML and statistical
modelling is the general goal of the analysis. Supervised ML is
predominantly concerned with obtaining the most accurate prediction
possible of the outcome variable. While the predictor variables are
necessary to obtain an accurate prediction, the specific relationship
between predictors and outcomes is often considered incidental. In
contrast, statistical modelling is primarily concerned with reliably
describing the relationship between certain predictor variables and the
outcome. While an accurate estimation model of the outcome is important,
it is only secondarily so as an indication of the reliability of the
model. Statistical analysis attempts to infer from the relationships in
the data analyzed patterns that can be generalized to the larger
population from which the data sample is drawn.

IML bridges the gap between the two modes of analysis by using
supervised ML techniques to describe the relationships between
predictors and the outcome even with models that are not intrinsically
interpretable. One major goal of such interpretation is to apply the
more flexible panoply of ML methods for the same goal as statistical
inference: reliably inferring relationships between predictors and the
outcome to a population larger than the sample. However, whereas
statistical inference is largely based on the interpretation of
coefficients of models that are variations of the general linear model
(GLM), most ML techniques do not have such coefficients. Thus, some
scholars have developed approaches to statistical inference specifically
tailored to ML.

Sechidis (2015) delves into the realm of hypothesis testing amid the
complexities of semi-supervised data, characterized by its partially
labelled and incomplete nature. His thesis meticulously examines the
unique challenges posed by this data type, particularly when drawing
conclusions from hypothesis tests. His exploration is not confined to
statistical significance alone; it also encompasses a thorough
consideration of effect sizes, thereby providing a more comprehensive
understanding of the analytical implications of the data.

Messner (2023) introduces some new measures with the ultimate goal of
enabling hypothesis testing from a model-agnostic approach using machine
learning. The key elements of his framework are:

\begin{itemize}
\tightlist
\item
  Determine the practical importance of effects based on the magnitude
  of \(f^2_v\): \textless{} 0.02 is trivial; 0.02 to 0.15 is small; 0.15
  to 0.35 is medium; and \textgreater= 0.35 is large. Crucially, rather
  than statistical significance, his effect-size-based hypothesis
  testing emphasizes the practical magnitude of effects.
\item
  Regardless of magnitude, determine if the effects are monotonic,
  applying the Mann-Kendall test for statistical significance.
\item
  For monotonic effects, determine if the effect is increasing or
  decreasing, applying the Theil-Sen test for statistical significance.
\end{itemize}

\hypertarget{opportunities-for-improvement}{%
\subsection{Opportunities for
improvement}\label{opportunities-for-improvement}}

Despite the various implementations and valuable extensions of the
original ALE concept, we see some ongoing issues that leave room for
improvement to better adapt it as a tool for statistical inference using
ML.

First, there is a crucial limitation of the standard data-only approach
to bootstrapping ALE values that we describe above (Flora 2023; Jumelle,
Kuhn-Regnier and Rajaratnam 2020). Ultimately, as an ML technique, ALE
values are usually calculated on a model to apply the explanations
obtained to understand future data. Even though, as we explain below,
the ALE should be calculated based on a final deployment model trained
on the full dataset, such a model needs to be developed by training it
on a training subset and evaluating it on a test subset. If a dataset is
too small for such a training-test split, then the model and dataset on
which ALE is calculated would not be representative of future data. In
this case, any generalizations that might be drawn from such analyses
strongly risk overfitting the data and there is no test set to qualify
such generalizations.

Second, while it is encouraging to see the development of various
model-agnostic effect size measures, those that exist tend to be limited
when the goal is to practically quantify the effects of variables. The
overall model MEC and the IAS (Molnar, Casalicchio and Bischl 2020) do
not show the effects for individual variables but only indicate
generally the average effects across all variables. While the
variable-specific version of the MEC attempts to address this concern,
it mainly indicates the extent to which a variable is non-linear but it
does not clearly quantify other important aspects of its possible
effect. Similarly, the impact score (Lötsch and Ultsch 2020) and
\(f^2_v\) (Messner 2023) indicate the overall effect of a predictor
variable on the outcome but do not shed much light on the nature of the
effect. The tests for monotonicity and slope (Messner 2023) are an
important step in the direction of clarifying the nature of the effect,
but by generalizing the overall trend in one tendency (e.g.,
monotonically increasing or non-monotonic), the details of complex
relationships might not be sufficiently clear.

A third concern is the computational practicality of some of the
measures. For example, Friedman's H is a useful measure for interaction
strength, but it is computationally rather expensive to calculate
(Molnar 2022).

In the following sections of the article, we address these outstanding
concerns.

\hypertarget{illustrative-datasets-and-models}{%
\section{Illustrative datasets and
models}\label{illustrative-datasets-and-models}}

Before we begin ALE analysis, in this section we describe the datasets
and the models that we will use for our illustrations. Because the
issues that we address are handled differently depending on the relative
size of the dataset, we work with two distinct datasets. We develop a
random forest model to analyze a moderately large dataset of diamond
prices and then a generalized additive model (GAM) to analyze a small
dataset of mathematics achievement scores.

\hypertarget{large-dataset-random-forest-model-for-diamond-prices}{%
\subsection{Large dataset: random forest model for diamond
prices}\label{large-dataset-random-forest-model-for-diamond-prices}}

For the analysis of a relatively large dataset, we use the
\texttt{diamonds} dataset, built-in with the \texttt{ggplot2} graphics
system (Wickham et al. 2023). We cleaned the original version by
removing duplicates (Mayer 2021) and invalid entries where the length
(x), width (y), or depth (z) is 0. Table~\ref{tbl-diamonds-dataset}
presents the description of the modified dataset. The outcome variable
that is the focus of our analysis is the \texttt{price} of a diamond,
with minimum \$326, median \$3,365, and maximum \$18,823 in USD.

\hypertarget{tbl-diamonds-dataset}{}
\begin{longtable}[]{@{}
  >{\raggedright\arraybackslash}p{(\columnwidth - 2\tabcolsep) * \real{0.1150}}
  >{\raggedright\arraybackslash}p{(\columnwidth - 2\tabcolsep) * \real{0.8850}}@{}}
\caption{\label{tbl-diamonds-dataset}Large dataset: diamond
prices}\tabularnewline
\toprule\noalign{}
\begin{minipage}[b]{\linewidth}\raggedright
Variable
\end{minipage} & \begin{minipage}[b]{\linewidth}\raggedright
Description
\end{minipage} \\
\midrule\noalign{}
\endfirsthead
\toprule\noalign{}
\begin{minipage}[b]{\linewidth}\raggedright
Variable
\end{minipage} & \begin{minipage}[b]{\linewidth}\raggedright
Description
\end{minipage} \\
\midrule\noalign{}
\endhead
\bottomrule\noalign{}
\endlastfoot
price & price in US dollars (\$326--\$18,823) \\
carat & weight of the diamond (0.2--5.01) \\
cut & quality of the cut (Fair, Good, Very Good, Premium, Ideal) \\
color & diamond color, from D (best) to J (worst) \\
clarity & a measurement of how clear the diamond is (I1 (worst), SI2,
SI1, VS2, VS1, VVS2, VVS1, IF (best)) \\
x\_length & length in mm (3.73--10.74) \\
y\_width & width in mm (3.68--58.9) \\
z\_depth & depth in mm (1.07--31.8) \\
depth\_pct & total depth percentage = z / mean(x, y) = 2 * z / (x + y)
(43--79) \\
table & width of top of diamond relative to widest point (43--95) \\
rand\_norm & a completely random variable (-4.9 to 4.4) \\
\end{longtable}

Of particular note is the variable \texttt{rand\_norm}. We have added
this completely random variable (with a normal distribution) to
demonstrate what randomness looks like in our analysis. (However, we
deliberately selected the specific random seed of 6 because it
highlights some particularly interesting points.)

There is considerable controversy regarding the appropriate data subset
for IML techniques such as ALE. For example, Molnar (2022) (Section
8.5.3) presents arguments in favour of alternatively analyzing the
training or the test subsets. In favour of the test set, he argues that
feature importance depends on model error estimates and error estimates
from training data are unreliable; so, the test set should be used. In
favour of the training subset, he argues that IML techniques like PDP
that reflect the relationship between predictors and outcomes do not
depend on model error, but rather, we want to use as much data as
possible to characterize the model, rather than being restricted by the
limited test data.

However, in contrast, we contend that the purpose of IML is not to
measure model error, which is the context where overfitting compels us
to split the data into distinct training and test subsets. Rather, IML
aims to characterize the entire dataset. It is beyond the scope of this
article to fully develop this argument, but the arguments for training
or test sets seem to be limited in their perspective to the model
development stages of ML and lose sight of the ultimate goal of ML:
model deployment. In ML operations (MLops), it is well established that
after training and evaluating a model, the final model that will be
deployed into production is one that is trained on the full dataset
using the algorithm and hyperparameters selected from the model
development stage. Because the goal of IML is to characterize the full
scenario based on the best information available, we argue that IML
techniques like ALE should be applied neither on training subsets nor on
test subsets but on a final deployment model after training and
evaluation. This final deployment should be trained on the full dataset
to give the best possible model for production deployment. So, this is
the model and dataset scope that we use for the rest of this article
when analyzing the large dataset.

ALE is a model-agnostic IML approach, that is, it works with any kind of
ML model. For this demonstration, we train a random forest to predict
diamond prices, selected to demonstrate the benefits of ALE to a
notoriously black-box model. To keep the article illustrations simple,
we assume that the hyperparameters we use were determined by proper
cross-validation of the data. Thus, the random forest that we train is a
final deployment model appropriate for IML analysis.

\hypertarget{small-dataset-generalized-additive-model-for-mathematics-achievement-scores}{%
\subsection{Small dataset: generalized additive model for mathematics
achievement
scores}\label{small-dataset-generalized-additive-model-for-mathematics-achievement-scores}}

We demonstrate ALE statistics on small datasets with a dataset composed
and transformed from the \texttt{nlme} package (Pinheiro et al. 2023).
The structure has 160 rows, each of which refers to a school whose
students have taken a mathematics achievement test.
Table~\ref{tbl-math-dataset} describes the data based on documentation
from the \texttt{nlme} package but, unfortunately, many details are not
quite clear.

\hypertarget{tbl-math-dataset}{}
\begin{longtable}[]{@{}
  >{\raggedright\arraybackslash}p{(\columnwidth - 4\tabcolsep) * \real{0.1604}}
  >{\raggedright\arraybackslash}p{(\columnwidth - 4\tabcolsep) * \real{0.0943}}
  >{\raggedright\arraybackslash}p{(\columnwidth - 4\tabcolsep) * \real{0.7358}}@{}}
\caption{\label{tbl-math-dataset}Small dataset: average mathematics
achievement scores for schools}\tabularnewline
\toprule\noalign{}
\begin{minipage}[b]{\linewidth}\raggedright
variable
\end{minipage} & \begin{minipage}[b]{\linewidth}\raggedright
format
\end{minipage} & \begin{minipage}[b]{\linewidth}\raggedright
description
\end{minipage} \\
\midrule\noalign{}
\endfirsthead
\toprule\noalign{}
\begin{minipage}[b]{\linewidth}\raggedright
variable
\end{minipage} & \begin{minipage}[b]{\linewidth}\raggedright
format
\end{minipage} & \begin{minipage}[b]{\linewidth}\raggedright
description
\end{minipage} \\
\midrule\noalign{}
\endhead
\bottomrule\noalign{}
\endlastfoot
math\_avg & double & average mathematics achievement scores of all
students in the school \\
size & double & the number of students in the school \\
public & logical & \begin{minipage}[t]{\linewidth}\raggedright
TRUE if the school is in the public sector;\\
FALSE if in the Catholic sector\strut
\end{minipage} \\
academic\_ratio & double & the percentage of students on the academic
track \\
female\_ratio & double & percentage of students in the school that are
female \\
mean\_ses & double & \begin{minipage}[t]{\linewidth}\raggedright
mean socioeconomic status for the students in the school\\
(measurement is not quite clear)\strut
\end{minipage} \\
minority\_ratio & double & percentage of students that are members of a
minority racial group \\
high\_minority & logical & \begin{minipage}[t]{\linewidth}\raggedright
TRUE if the school has a high ratio of students of minority racial
groups\\
(unclear, but perhaps relative to the location of the school)\strut
\end{minipage} \\
discrim & double & \begin{minipage}[t]{\linewidth}\raggedright
the ``discrimination climate''\\
(perhaps an indication of extent of racial discrimination in the
school?)\strut
\end{minipage} \\
rand\_norm & double & a completely random variable \\
\end{longtable}

Of particular note again is the variable \texttt{rand\_norm}. As with
the large dataset, we have added this completely random variable (with a
normal distribution) to demonstrate what randomness looks like in our
analysis. (However, we selected the specific random seed of 6 because it
highlights some particularly interesting points.)

The outcome variable that is the focus of our analysis is
\texttt{math\_avg}, the average mathematics achievement scores of all
students in each school, with minimum 4.24, median 12.9, and maximum
19.7.

Because the samples here are relatively small, we will use general
additive models (GAM) for the modelling (Wood 2023). GAM is an extension
of statistical regression analysis that lets the model fit flexible
patterns in the data instead of being restricted to the best-fitting
straight line. It is an ideal approach for samples that are too small
for machine learning because it provides flexible curves unlike ordinary
least squares regression yet will not overfit excessively as would most
machine learning techniques when working with such small samples (Ross
2019).

\hypertarget{bootstrapping-of-accumulated-local-effects}{%
\section{Bootstrapping of accumulated local
effects}\label{bootstrapping-of-accumulated-local-effects}}

We now come to the application of ALE to address the issues we
highlighted above in the introduction and related work sections. In this
section, we explore two distinct bootstrapping approaches, each applied
to datasets of varying sizes and characteristics. Initially, we begin
with the data-only bootstrap approach to improve the reliability of ALE
plots by providing confidence intervals that offer more reliable
estimates than those derived from a single ALE calculation. Next, we
assess the imperative of model bootstrapping to mitigate the risk of
overfitting in the context of smaller datasets.

\hypertarget{data-only-bootstrapping-of-a-large-dataset-diamonds}{%
\subsection{Data-only bootstrapping of a large dataset
(diamonds)}\label{data-only-bootstrapping-of-a-large-dataset-diamonds}}

\begin{figure}

{\centering \includegraphics{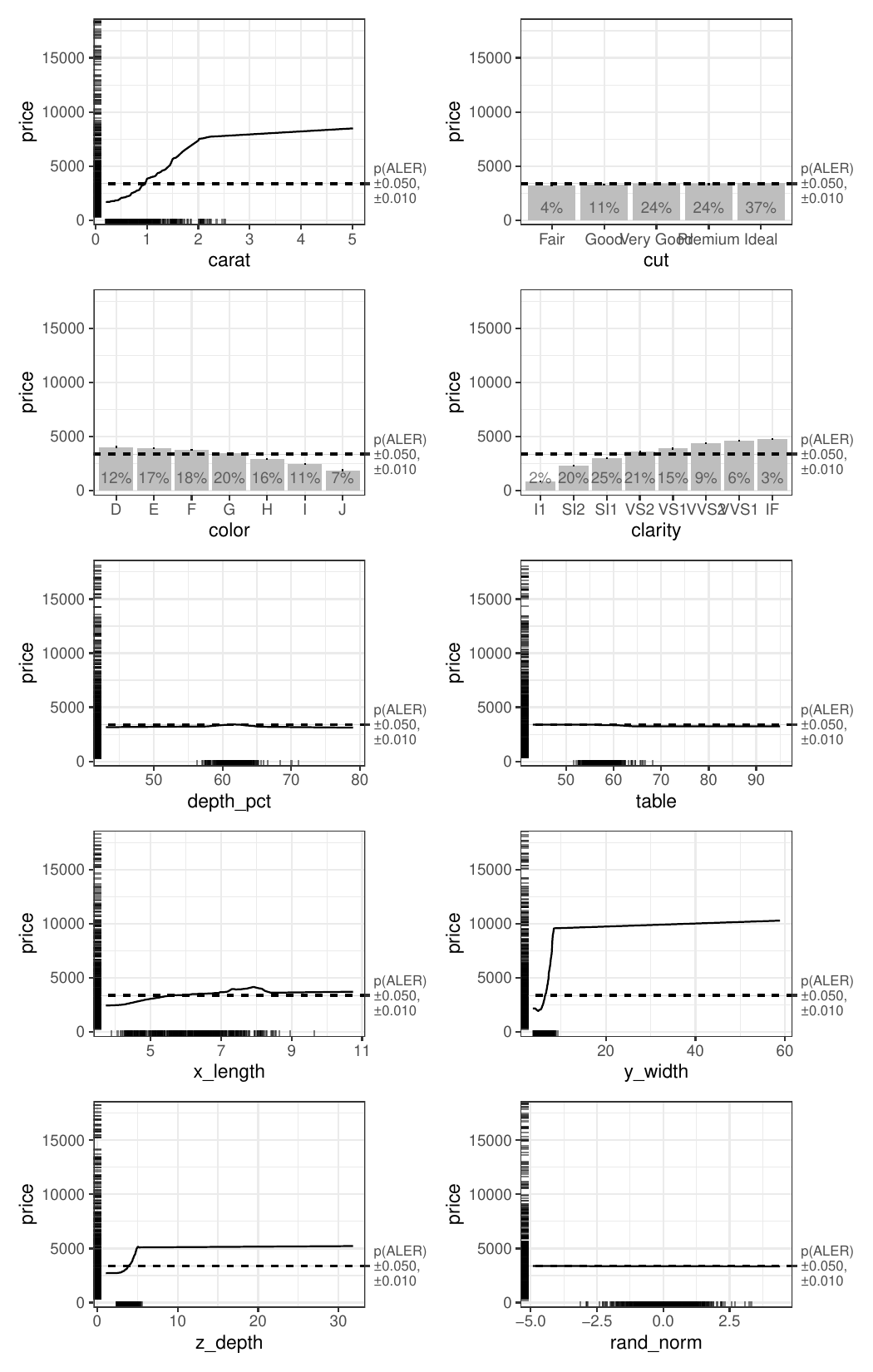}

}

\caption{\label{fig-large_ale_simple}Simple ALE plots for random forest
model of diamond prices}

\end{figure}

Figure~\ref{fig-large_ale_simple} displays the simple ALE plots for the
random forest model of diamond prices. (We refer to these plots as
``simple'' because they simply calculate ALE without confidence
intervals.) In this case and all our other analyses on the
\texttt{diamonds} dataset, the ALE is calculated on the full dataset
using the final deployment random forest model.

\begin{figure}

{\centering \includegraphics{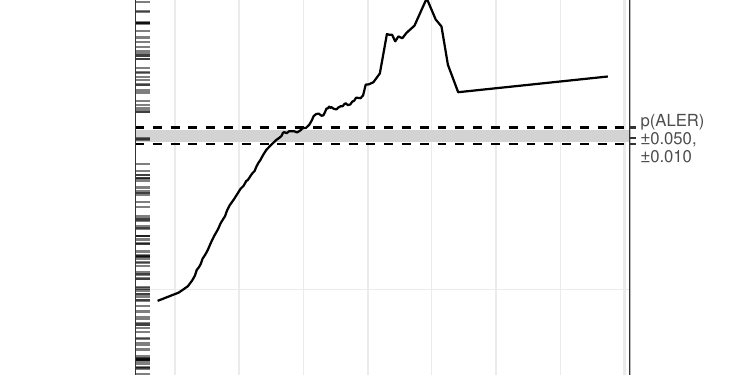}

}

\caption{\label{fig-zoom_x_length}Zoom-in of ALE plot for x\_length for
diamond prices}

\end{figure}

A significant feature of the \texttt{ale} package that we use in this
article is that it centres ALE values on the median, unlike the original
\texttt{ALEPlot} implementation (Apley 2018), which centres them on
zero. Although the \texttt{ale} package lets the user centre on zero or
even on the mean, the median is particularly crucial for the statistical
inference framework that the \texttt{ale} package offers. An essential
element of its visualization is the middle grey band. Because the range
of \texttt{price} values in Figure~\ref{fig-large_ale_simple} is so
broad, we zoom in on the \texttt{x\_length} variable in
Figure~\ref{fig-zoom_x_length} to see this more clearly. Below, we
explain what exactly the ALE range (ALER) means, but for now, we can
briefly explain a few elements.

The approximate middle of the grey band is the median of the \emph{y}
outcome variables in the dataset (\texttt{price}, in our case). The
middle tick on the right y-axis indicates the exact median. We call this
grey band the ``ALER band''. 95\% of random variables had ALE values
that fully lay within the ALER band. The dashed lines above and below
the ALER band expand the boundaries to where 99\% of the random
variables were constrained. These boundaries could be considered as
demarcating an extended or outward ALER band.

\begin{figure}

{\centering \includegraphics{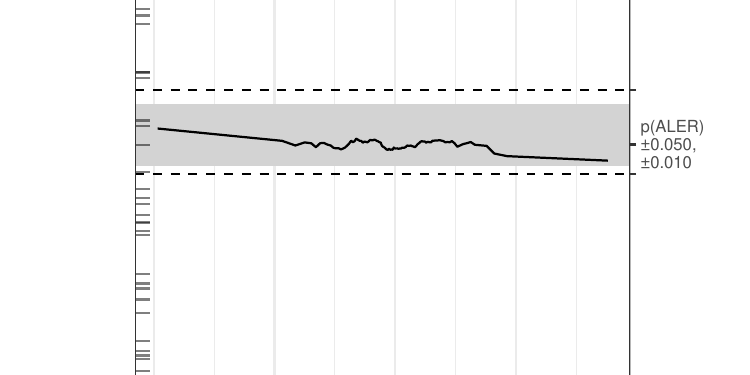}

}

\caption{\label{fig-zoom_rand_norm}Zoom-in of ALE plot for rand\_norm
for diamond prices}

\end{figure}

The idea is that if the ALE values of any predictor variable falls fully
within the ALER band, then it has no greater effect than 95\% of purely
random variables. Moreover, to consider any effect in the ALE plot to be
statistically significant (that is, non-random), there should be no
overlap between the bootstrapped confidence regions of a predictor
variable and the ALER band. As we can see in
Figure~\ref{fig-zoom_rand_norm} where we zoom in on the random variable,
\texttt{rand\_norm} fully falls within the ALER band, indicating the
randomness of its effect.

Looking at the simple ALE plots in Figure~\ref{fig-large_ale_simple}, we
could draw the following initial conclusions:

\textbf{Verify based on p\_funs}

\begin{itemize}
\tightlist
\item
  The prices increase sharply with the \texttt{carat} of the diamonds
  until they plateau at the highest end of the domain just above 2
  carat. The rug plot indicates that there are very few diamonds with a
  higher carat than that.
\item
  \texttt{cut} does not seem to have much of an effect on prices.
\item
  The D, E, and F \texttt{color}s have higher prices on average; H, I,
  and J have lower prices.
\item
  The I1, S12, and SI1 \texttt{clarity} categories have lower prices
  while VVS2, VVS1, and IF have higher prices. VS2 and VS1 do not seem
  to have as much of an effect.
\item
  \texttt{depth\_pct} and \texttt{table} do not have much of an effect
  on prices.
\item
  \texttt{x\_length} has a very gradual increasing effect on price
  throughout its domain.
\item
  As \texttt{y\_width} increases, \texttt{price} sharply increases with
  it. Although the plot shows a long plateau, the rug plot indicates
  that the plateau is predominantly in regions of high \texttt{y\_width}
  where there are few diamonds.
\item
  \texttt{z\_depth} has a similar pattern to that of \texttt{y\_width},
  though its effect is not as extreme.
\item
  As we have noted, the random variable \texttt{rand\_norm} does not
  have much of an effect on prices. We will continue pay attention to
  this variable in our subsequent analyses.
\end{itemize}

Despite these interesting initial results, it is crucial to bootstrap
the ALE results to ensure that they are reliable, that is, generalizable
to data beyond the sample on which the model was built. So, now we
recreate bootstrapped versions of the ALE data and plots. Specifically,
because we assume that the random forest model is the product of
hyperparameter optimization on properly validated subsets of data,
recalculate ALE using the same final deployment model on each bootstrap
sample; this is data-only bootstrapping.

\begin{figure}

{\centering \includegraphics{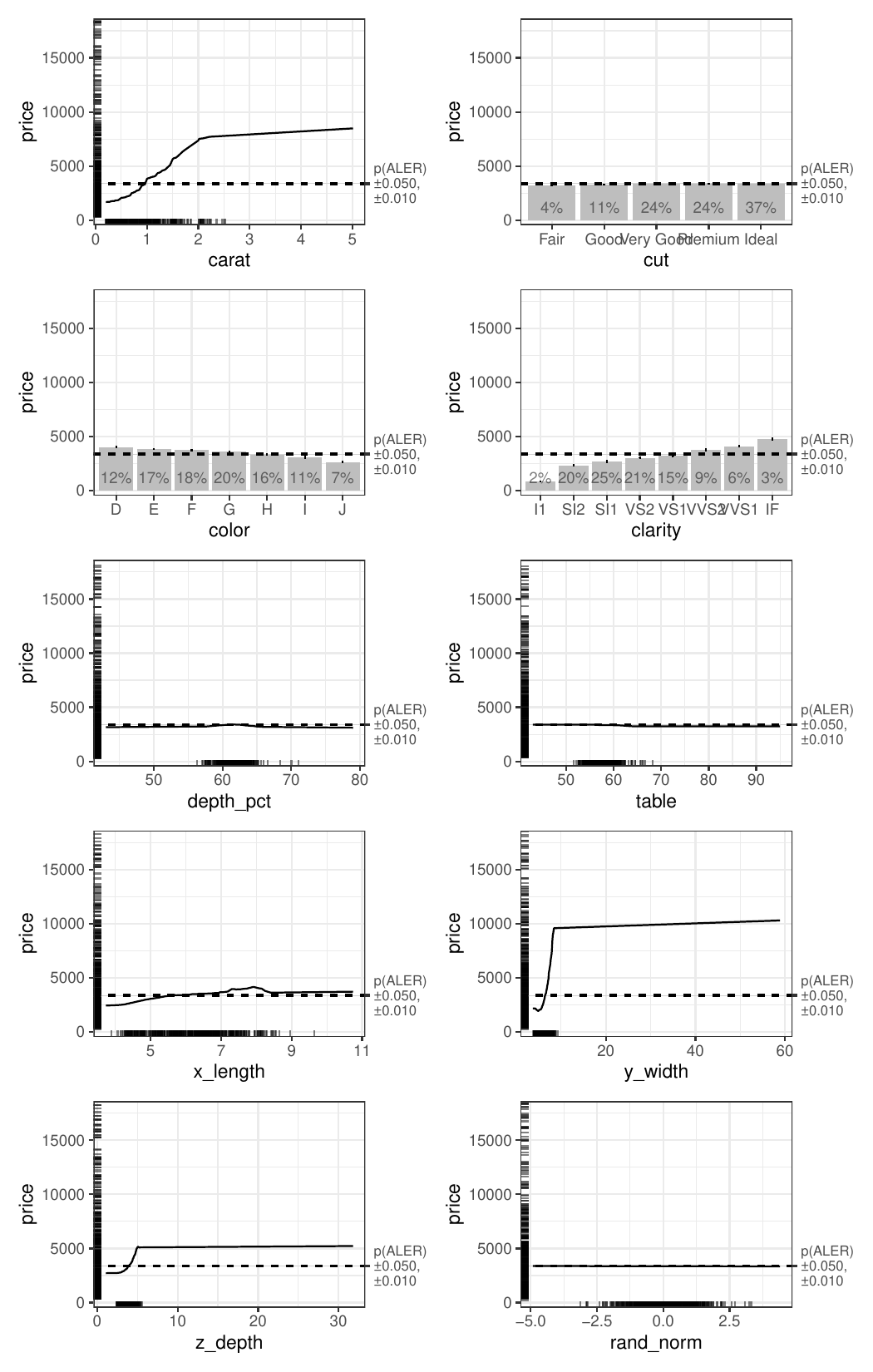}

}

\caption{\label{fig-large_ale}Bootstrapped ALE plots for random forest
model of diamond prices}

\end{figure}

ALE is a relatively stable IML algorithm (compared to others like PDP),
so 100 bootstrap samples should be sufficient for relatively stable
results, especially for model development (Apley and Zhu 2020).

The bootstrapped results in Figure~\ref{fig-large_ale} are very similar
to single (non-bootstrapped) ALE results from
Figure~\ref{fig-large_ale_simple}, which attests to the stability of ALE
results when working with moderately large datasets. However,
bootstrapping adds confidence intervals to the plot (we use a 95\%
interval in our illustrations), which adds some uncertainty when certain
values are close to the ALER band as to whether they overlap the band or
not. We present tools to clarify this point below.

\hypertarget{model-bootstrapping-of-a-small-dataset-math-achievement}{%
\subsection{Model bootstrapping of a small dataset (math
achievement)}\label{model-bootstrapping-of-a-small-dataset-math-achievement}}

\begin{figure}

{\centering \includegraphics{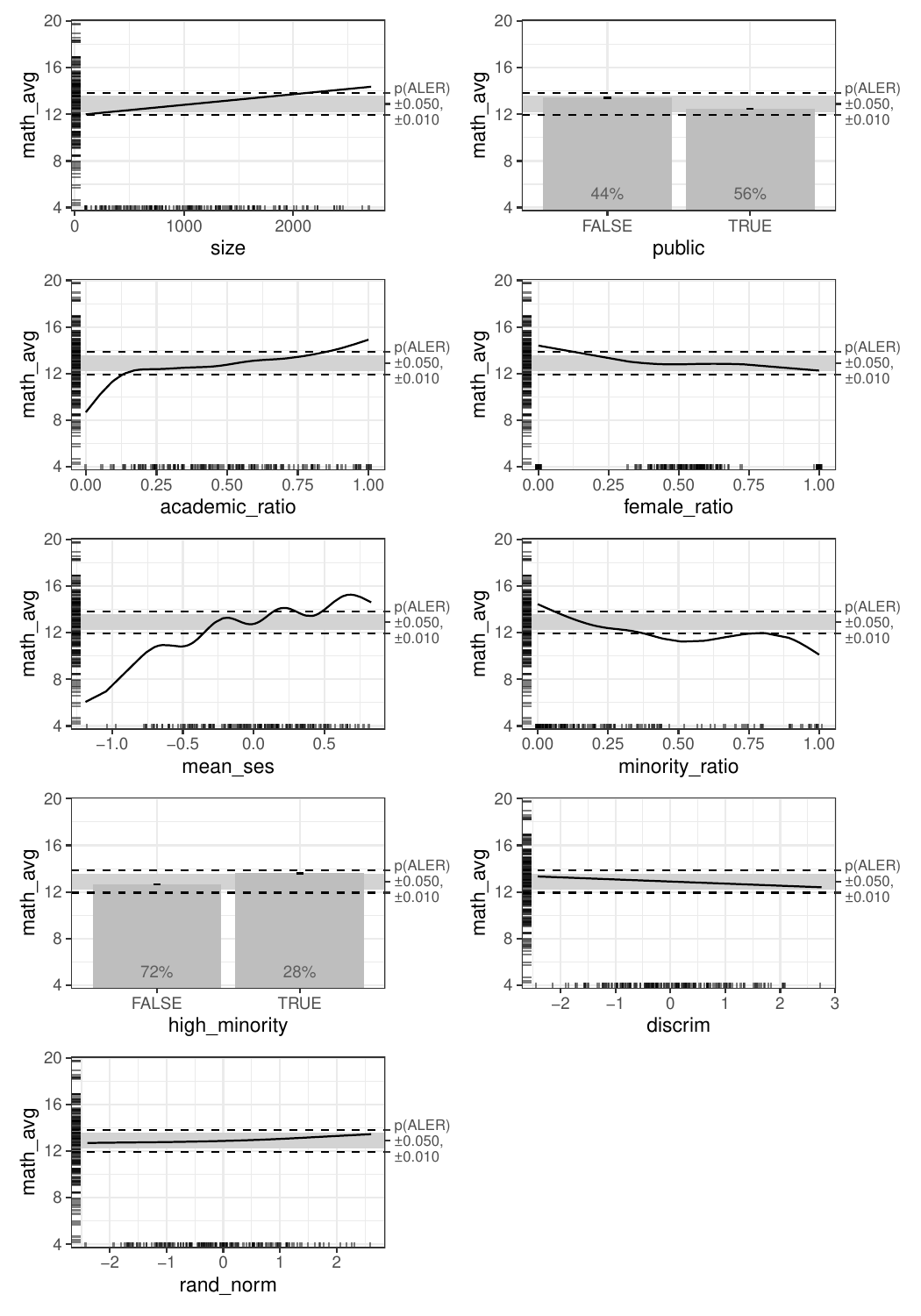}

}

\caption{\label{fig-small_ale_simple}Simple ALE plots for GAM of
mathematics achievement scores}

\end{figure}

Turning to the case of small datasets, we again begin with simple ALE
plots to see what results look like without bootstrapping. With this
smaller dataset, the ALER band (representing the range of 95\% of random
variables) is much wider and we can more easily see each variables'
effects in relation to it. The ALE plots in
Figure~\ref{fig-small_ale_simple} suggest that:

\begin{itemize}
\tightlist
\item
  \texttt{size}, \texttt{academic\_ratio}, and \texttt{mean\_ses} have
  positive relationships with math achievement scores. The relationship
  with \texttt{size} is linear, but those with the other variables is
  non-linear.
\item
  \texttt{female\_ratio} and \texttt{minority\_ratio} seem to have
  negative relationships with math achievement scores. The negative
  relationship of the \texttt{female\_ratio} does not seem to be that
  strong, though it seems to be driven by the fact that all-male schools
  have above-average scores while all-female schools have below-average
  scores.
\item
  \texttt{public} schools seem to have lower scores on average while
  those with \texttt{high\_minority} ratios have higher scores, but both
  values fall within the ALER band. Thus, the apparent effect might not
  be meaningful.
\item
  The discrimination climate \texttt{discrim}, though somewhat negative,
  falls fully within the ALER band; so it does not seem to have a
  meaningful effect.
\item
  The random variable \texttt{rand\_norm}, though somewhat positive,
  also falls fully within the ALER band.
\end{itemize}

However, before reading too much into these results, we must remember
that results that are not bootstrapped are simply not reliable.

\hypertarget{inappropriate-data-only-bootstrapping-of-a-small-dataset}{%
\subsubsection{Inappropriate data-only bootstrapping of a small
dataset}\label{inappropriate-data-only-bootstrapping-of-a-small-dataset}}

\begin{figure}

{\centering \includegraphics{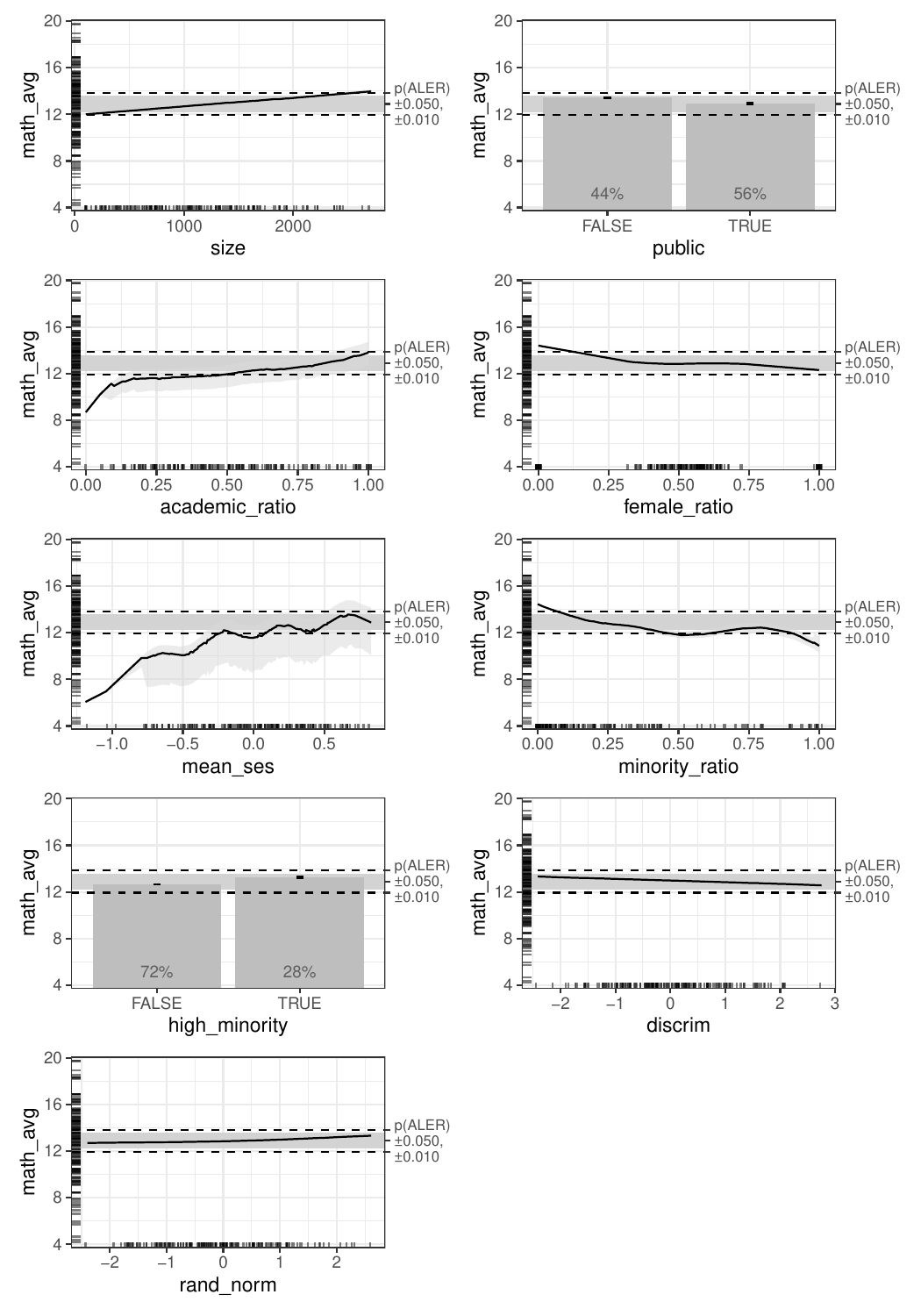}

}

\caption{\label{fig-small_ale_boot}Data-only (inappropriate)
bootstrapped ALE plots for GAM of mathematics achievement scores}

\end{figure}

To illustrate the implications of the different bootstrap approaches, we
will first carry out a data-only bootstrap as we did with the 100
bootstrap iterations

With data-only bootstrapping, the results in
Figure~\ref{fig-small_ale_boot} are very similar to those without
bootstrapping in Figure~\ref{fig-small_ale_simple}, with only
\texttt{mean\_ses} displaying wide confidence bands that suggest that
some of the ranges of its values might not be relevant. However,
data-only bootstrapping is not the appropriate approach for small
samples. The issue is not directly the size of the dataset but the fact
that the model was trained on the entire dataset rather than on a sample
distinct from that on which the ALE data and plots are created. (In
contrast, we assume that our large dataset example was trained on a
deployment model developed by prior cross-validation.) Thus, for the
results to represent a broader sample than that of the small dataset, we
must not only bootstrap the entire dataset but we must also retrain the
entire model on each of the bootstrapped datasets. Although the problem
might not be so readily evident by examining the inappropriately
bootstrapped plot above, we revisit this issue below when we discuss
bootstrap-based inference with ALE and we see how inappropriate this
approach is in handling the random variable \texttt{rand\_norm}.

\hypertarget{appropriate-full-model-bootstrapping-of-a-small-dataset}{%
\subsubsection{Appropriate full-model bootstrapping of a small
dataset}\label{appropriate-full-model-bootstrapping-of-a-small-dataset}}

Data-only bootstrapping is not appropriate for small datasets where a
model is developed without proper validation of its hyperparameters
because there is a high risk of overfitting. Indeed, This is one of the
classic cases that the bootstrap was originally designed to address. In
such cases, the entire dataset should be bootstrapped and the model
should be retrained on each bootstrap sample. Thus, there will be as
many models as there are bootstrap samples. Any necessary calculations
(typically overall model statistics and variable coefficients) are
calculated for each model and the averages, quantiles, etc. of these
calculations across model-bootstrap samples are calculated as the
bootstrapped estimates.

We call this approach \textbf{model bootstrapping} because not only is
the data bootstrapped, but the model itself is also bootstrapped. This
might also be called model-dependent bootstrap or resampling with model
re-fitting. Whereas data-only bootstrapping has \(n_{it}\) (number of
iterations) bootstrap samples of the data but only one model, model
bootstrapping has \(n_{it}\) distinct models, one for each of the
\(n_{it}\) bootstrap samples. Model bootstrapping is much slower than
data-only bootstrapping because the model has to be retrained \(n_{it}\)
times. While model bootstrapping does not mitigate overfitting, it
provides confidence intervals that effectively give more reliable
estimates than naively accepting only the single estimates from a single
model on the entire dataset. It should be considered mandatory for
datasets that are too small to be split further into distinct training
and test sets.

Model bootstrapping for ALE is similar to data-only bootstrapping for
ALE in that fixed ALE intervals must be calculated from the full dataset
and then applied to each bootstrap sample, even though in this case each
bootstrap sample has a different model. The essential difference is that
a different model is used to calculate the ALE values each time---the
model specific to each bootstrap sample.

Thus, we apply the appropriate model bootstrapping approach to the math
dataset. Model bootstrapping is particularly slow, even on small
datasets, since the entire process is repeated that many times. However,
100 iterations should be sufficiently stable for model building.

\begin{figure}

{\centering \includegraphics{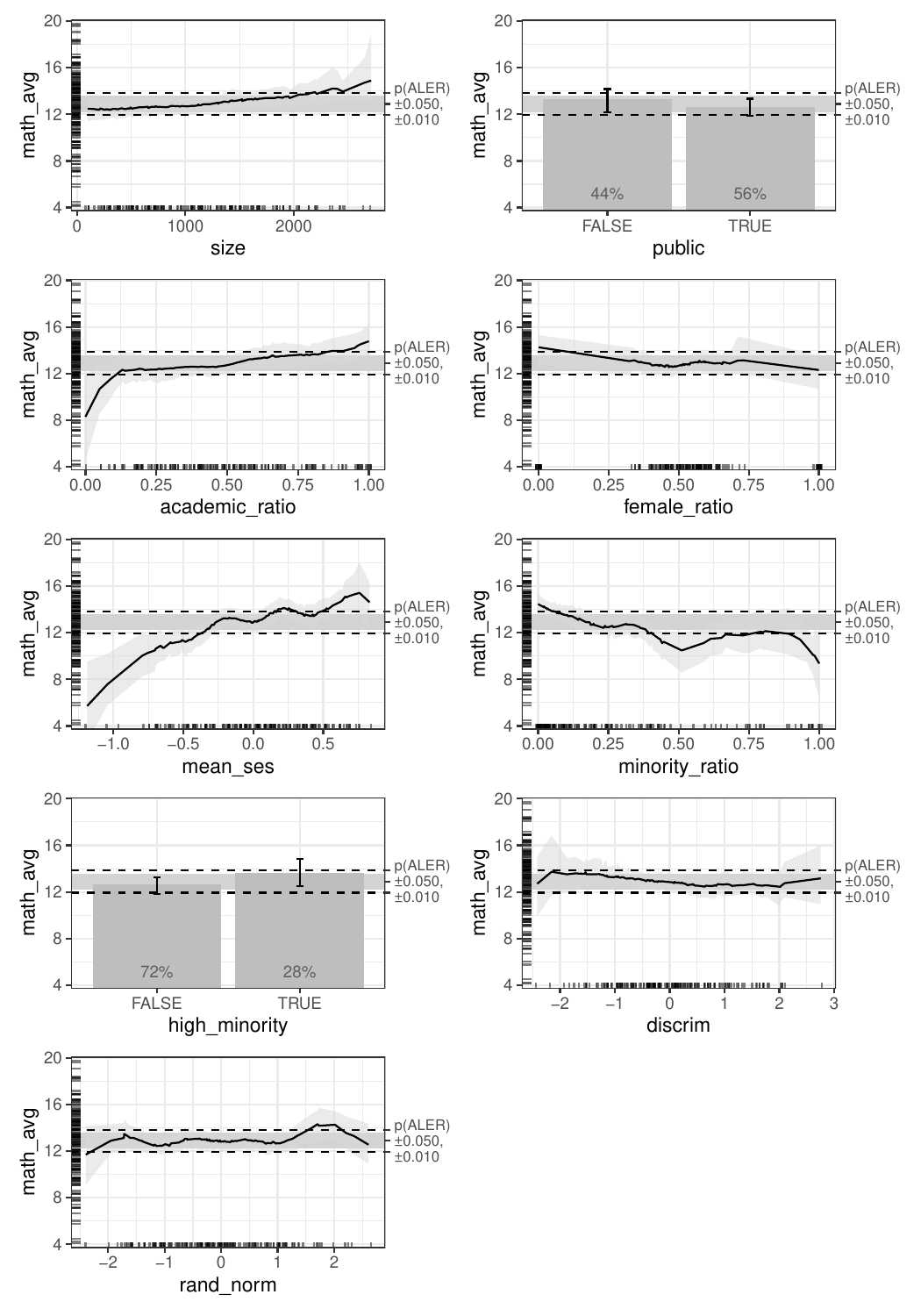}

}

\caption{\label{fig-small_ale}Model (appropriate) bootstrapped ALE plots
for GAM of mathematics achievement scores}

\end{figure}

With model bootstrapping, although the general patterns of relationships
in Figure~\ref{fig-small_ale} are similar to those of the data-only
bootstrapping in Figure~\ref{fig-small_ale_boot}, the confidence bands
are generally much wider, which indicates that the data-only bootstrap
greatly exaggerated the reliability of the ALE results for this small
sample. We can see that most variables seem to have some sort of mean
effect across various values. However, for statistical inference, our
focus must be on the bootstrap intervals. For an effect to be considered
statistically significant, there should be no overlap between the
confidence regions of a predictor variable and the ALER band.

For categorical variables (\texttt{public} and \texttt{high\_minority}
above), the confidence interval bands for all categories overlap the
ALER band. The confidence interval bands indicate two useful pieces of
information to us. When we compare them to the ALER band, their overlap
or lack thereof tells us about the practical significance of the
category. When we compare the confidence bands of one category with
those of others, it allows us to assess if the category has a
statistically significant effect that is different from that of the
other categories; this is equivalent to the regular interpretation of
coefficients for GAM and other GLM models. In both cases, the confidence
interval bands of the \texttt{TRUE} and \texttt{FALSE} categories
overlap each other, indicating that there is no statistically
significant difference between categories. Each confidence interval band
overlaps the ALER band, indicating that none of the effects is
practically significant, either.

For numeric variables, the confidence regions overlap the ALER band for
most of the domains of the predictor variables except for some regions
that we will examine. The extreme points of each variable (except for
\texttt{discrim} and \texttt{female\_ratio}) are usually either slightly
below or slightly above the ALER band, indicating that extreme values
have the most extreme effects: math achievement increases with
increasing school size, academic track ratio, and mean socioeconomic
status, whereas it decreases with increasing minority ratio. The ratio
of females and the discrimination climate both overlap the ALER band for
the entirety of their domains, so any apparent trends are not supported
by the data.

Of particular interest is the random variable \texttt{rand\_norm}, whose
average bootstrapped ALE appears to show some sort of pattern. However,
we note that the 95\% confidence intervals we use mean that if we were
to retry the analysis for twenty different random seeds, we would expect
at least one of the random variables to partially escape the bounds of
the 5\% ALER band. We will return below to the implications of random
variables in ALE analysis.

\hypertarget{ale-effect-size-measures}{%
\section{ALE effect size measures}\label{ale-effect-size-measures}}

Now that we can calculate reliable ALE effects by bootstrapping, we
proceed to consider the strengths of the variable effects. In all cases,
we focus on the average of the ALE values calculated from the bootstrap
samples. The ``average'' here is usually the mean, though the median can
also be used.

In general, effect size measures are most frequently used in statistical
analysis, which typically analyzes relatively small datasets. They are
also used to a lesser extent in machine learning with large datasets,
typically in the context of variable importance measures for IML (Molnar
2022). Thus, the measures that we describe here are relevant to both
small and large datasets. Given their relatively greater relevance for
statistical analysis of smaller datasets, we will demonstrate them in
this section almost exclusively using the mathematics achievement
dataset. However, we note that everything we describe here is fully
relevant to large datasets; indeed, we end this section with a pertinent
illustration

\hypertarget{ale-effect-size-plot}{%
\subsection{ALE effect size plot}\label{ale-effect-size-plot}}

\begin{figure}

{\centering \includegraphics{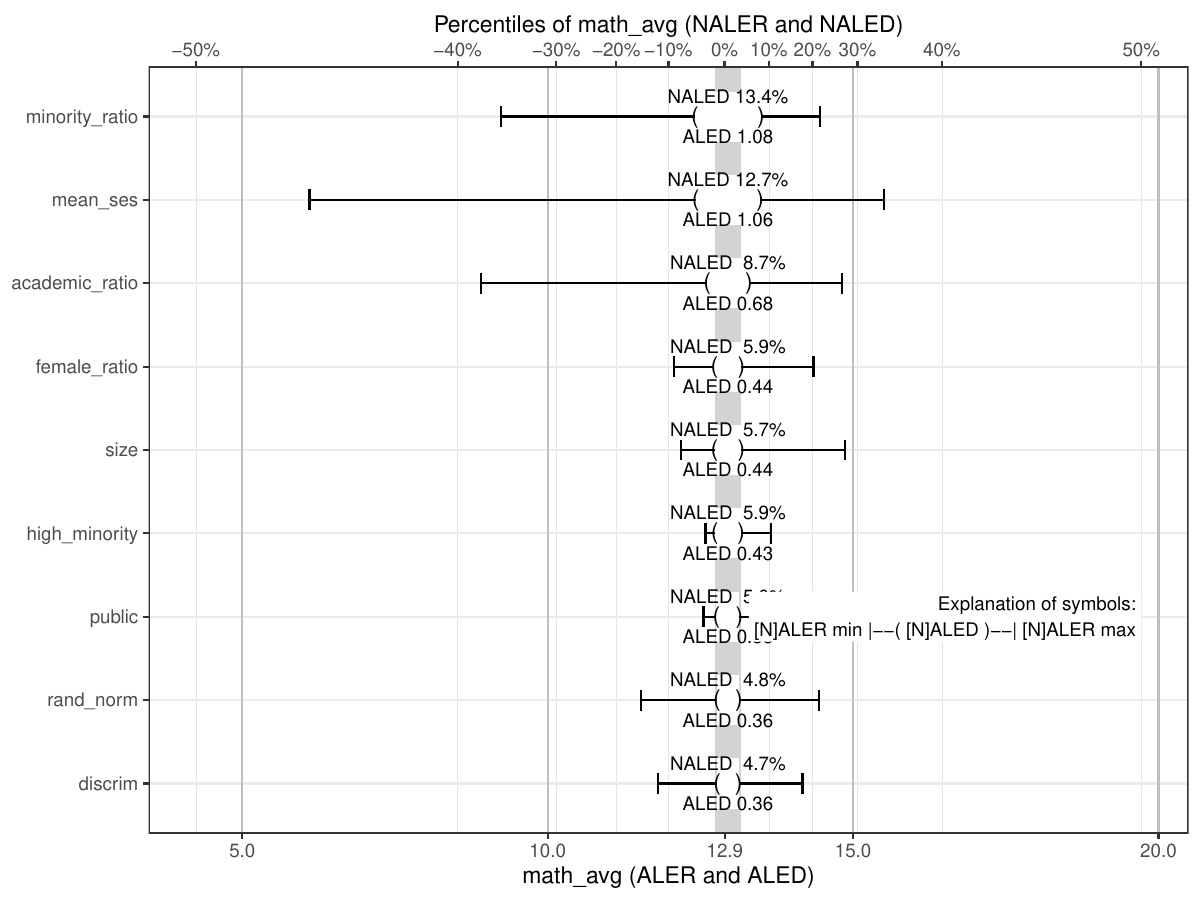}

}

\caption{\label{fig-math_effects}ALE effects plot for mathematics
achievement scores}

\end{figure}

Although ALE plots allow rapid and intuitive conclusions for statistical
inference, it is often helpful to have summary numbers that quantify the
average strengths of the effects of a variable. Thus, we have developed
a collection of effect size measures based on ALE tailored for intuitive
interpretation. To understand the intuition underlying the various ALE
effect size measures, even before we explain the measures in detail, it
is useful to first examine the \textbf{ALE effects plot} in
Figure~\ref{fig-math_effects} that graphically summarizes the effect
sizes of all the variables in the ALE analysis.

This plot requires some explanation:

\begin{itemize}
\tightlist
\item
  The y- (vertical) axis displays the \emph{x} variables, rather than
  the x-axis. This is consistent with most effect size plots because
  they list the full names of variables. It is more readable to list
  them as labels on the y-axis than the other way around.
\item
  The x- (horizontal) axis thus displays the \emph{y} (outcome)
  variable. But there are two representations of this same axis, one at
  the bottom and one at the top.
\item
  On the bottom is a more typical axis of the outcome variable, in our
  case, \texttt{math\_avg}. It is scaled as expected. In our case, the
  axis breaks default to five units each from 5 to 20, evenly spaced.
  The median of 13 is also specifically marked.
\item
  On the top, the outcome variable is expressed as percentiles ranging
  from 0\% (the minimum outcome value in the dataset) to 100\% (the
  maximum). It is divided into ten deciles of 10\% each. Because
  percentiles are usually not evenly distributed in a dataset, the
  decile breaks are not evenly spaced.
\item
  Thus, this plot has two x-axes, the lower one in units of the outcome
  variable and the upper one in percentiles of the outcome variable. To
  reduce the confusion, the major vertical grid lines that are slightly
  darker align with the units of the outcome (lower axis) and the minor
  vertical grid lines that are slightly lighter align with the
  percentiles (upper axis).
\item
  The vertical grey band in the middle is the NALED band. Its width is
  the 0.05 p-value of the NALED (explained below). That is, 95\% of
  random variables had a NALED equal or smaller than that width.
\item
  The variables on the horizontal axis are sorted by decreasing ALED and
  then NALED value (explained below).
\end{itemize}

Although it is somewhat confusing to have two axes, the percentiles are
a direct transformation of the raw outcome values. The first two base
ALE effect size measures below are in units of the outcome variable
while their normalized versions are in percentiles of the outcome. Thus,
the same plot can display the two kinds of measures simultaneously.
Referring to this plot can help to understand each of the measures,
which we proceed to explain in detail.

\hypertarget{ale-effect-size-measures-on-the-scale-of-the-y-outcome-variable}{%
\subsection{\texorpdfstring{ALE effect size measures on the scale of the
\emph{y} outcome
variable}{ALE effect size measures on the scale of the y outcome variable}}\label{ale-effect-size-measures-on-the-scale-of-the-y-outcome-variable}}

In this subsection, we describe the two base effect size measures that
are scaled on the same unit of measure as the \emph{y} outcome variable.

However, before we explain these measures in detail, we must reiterate
the timeless reminder that correlation is not causation. So, none of the
scores necessarily means that an \emph{x} variable \emph{causes} a
certain effect on the \emph{y} outcome; we can only say that the ALE
effect size measures indicate associated or related variations between
the two variables.

\hypertarget{ale-range-aler}{%
\subsubsection{ALE range (ALER)}\label{ale-range-aler}}

The easiest ALE statistic to understand is the ALE range (ALER), so we
begin there. It is simply the range from the minimum to the maximum of
any \texttt{ale\_y} value for that variable. Mathematically, we see this
in Equation~\ref{eq-aler}.

\begin{equation}\protect\hypertarget{eq-aler}{}{
\mathrm{ALER}(\mathrm{ale\_y}) = \{ \min(\mathrm{ale\_y}), \max(\mathrm{ale\_y}) \}
}\label{eq-aler}\end{equation}

where \(\mathrm{ale\_y}\) is the vector of ALE \emph{y} values for a
variable.

We note that all the ALE effect size measures are centred on zero so
that they are consistent regardless of whether the ALE plots are centred
on zero (as in the original \texttt{ALEPlot} implementation (Apley
2018)) or on the median (as in the default with the \texttt{ale} package
that we use in this article). Specifically,

\begin{itemize}
\tightlist
\item
  \texttt{aler\_min}: minimum of any \texttt{ale\_y} value for the
  variable.
\item
  \texttt{aler\_max}: maximum of any \texttt{ale\_y} value for the
  variable.
\end{itemize}

ALER shows the extreme values of a variable's effect on the outcome. In
the effects plot above, it is indicated by the extreme ends of the
horizontal lines for each variable. We can access ALE effect size
measures through the \texttt{ale\$stats} element of the bootstrap result
object, with multiple views. To focus on all the measures for a specific
variable, we can access the \texttt{ale\$stats\$by\_term} element. In
Table~\ref{tbl-math_ale_stats_public}, we see the effect size measures
for the categorical \texttt{public}.

\hypertarget{tbl-math_ale_stats_public}{}
\begin{longtable}[t]{>{}lrrrrrr}
\caption{\label{tbl-math_ale_stats_public}Effect size measures for the categorical \texttt{public} from the math
achievements dataset }\tabularnewline

\toprule
statistic & estimate & p.value & conf.low & median & mean & conf.high\\
\midrule
\textbf{aled} & 0.376 & 0.000 & 0.019 & 0.332 & 0.376 & 0.968\\
\textbf{aler\_min} & -0.344 & 0.349 & -0.846 & -0.320 & -0.344 & -0.020\\
\textbf{aler\_max} & 0.421 & 0.245 & 0.017 & 0.369 & 0.421 & 1.198\\
\textbf{naled} & 5.029 & 0.004 & 0.000 & 4.720 & 5.029 & 11.785\\
\textbf{naler\_min} & -4.261 & 0.650 & -11.204 & -3.414 & -4.261 & 0.000\\
\addlinespace
\textbf{naler\_max} & 6.069 & 0.405 & 0.000 & 5.625 & 6.069 & 17.828\\
\bottomrule
\end{longtable}

We see there that \texttt{public} has an ALER of {[}-0.34, 0.42{]}. When
we consider that the median math score in the dataset is 12.9, this ALER
indicates that the minimum of any ALE \emph{y} value for \texttt{public}
(when \texttt{public\ ==\ TRUE}) is -0.34 below the median. This is
shown at the 12.6 mark in the plot above. The maximum
(\texttt{public\ ==\ FALSE}) is 0.42 above the median, shown at the 13.3
point above.

The unit for ALER is the same as the outcome variable; in our case, that
is \texttt{math\_avg} ranging from 2 to 20. No matter what the average
ALE values might be, the ALER quickly shows the minimum and maximum
effects of any value of the \emph{x} variable on the \emph{y} variable.

For contrast, in Table~\ref{tbl-math_ale_stats_mean_ses}, we see the ALE
effect size measures for the numeric \texttt{mean\_ses}.

\hypertarget{tbl-math_ale_stats_mean_ses}{}
\begin{longtable}[t]{>{}lrrrrrr}
\caption{\label{tbl-math_ale_stats_mean_ses}Effect size measures for the numeric \texttt{mean\_ses} from the math
achievements dataset }\tabularnewline

\toprule
statistic & estimate & p.value & conf.low & median & mean & conf.high\\
\midrule
\textbf{aled} & 1.055 & 0 & 0.610 & 1.053 & 1.055 & 1.446\\
\textbf{aler\_min} & -6.798 & 0 & -10.149 & -6.916 & -6.798 & -3.270\\
\textbf{aler\_max} & 2.604 & 0 & 1.283 & 2.528 & 2.604 & 4.860\\
\textbf{naled} & 12.705 & 0 & 7.959 & 12.528 & 12.705 & 17.317\\
\textbf{naler\_min} & -45.572 & 0 & -50.000 & -47.226 & -45.572 & -33.063\\
\addlinespace
\textbf{naler\_max} & 32.663 & 0 & 15.415 & 33.441 & 32.663 & 44.411\\
\bottomrule
\end{longtable}

The ALER for \texttt{mean\_ses} is considerably broader with -6.8 below
and 2.6 above the median.

\hypertarget{ale-deviation-aled}{%
\subsubsection{ALE deviation (ALED)}\label{ale-deviation-aled}}

While the ALE range shows the most extreme effects a variable might have
on the outcome, the ALE deviation indicates its average effect over its
full domain of values. The zero-centred ALE values, it is conceptually
similar to the weighted mean absolute error (MAE) of the ALE \emph{y}
values. Mathematically, we see this in Equation~\ref{eq-aled}.

\begin{equation}\protect\hypertarget{eq-aled}{}{
\mathrm{ALED}(\mathrm{ale\_y}, \mathrm{ale\_n}) = \frac{\sum_{i=1}^{k} \left| \mathrm{ale\_y}_i \times \mathrm{ale\_n}_i \right|}{\sum_{i=1}^{k} \mathrm{ale\_n}_i}
}\label{eq-aled}\end{equation}

where \(i\) is the index of \(k\) ALE \emph{x} intervals for the
variable (for a categorical variable, this is the number of distinct
categories), \(\mathrm{ale\_y}_i\) is the ALE \emph{y} value for the
\(i\)th ALE \emph{x} interval, and \(\mathrm{ale\_n}_i\) is the number
of rows of data in the \(i\)th ALE \emph{x} interval.

Based on its ALED, we can say that the average effect on math scores of
whether a school is in the public or Catholic sector is 0.38 (again, out
of a range from 2 to 20). In the effects plot above, the ALED is
indicated by a white box bounded by parentheses ( and ). As it is
centred on the median, we can readily see that the average effect of the
school sector barely exceeds the limits of the ALER band, indicating
that it barely exceeds our threshold of practical relevance. The average
effect for mean socioeconomic status is slightly higher at 1.06. We can
see on the plot that it slightly exceeds the ALER band on both sides,
indicating its slightly stronger effect. We will comment on the values
of other variables when we discuss the normalized versions of these
scores, to which we proceed next.

\hypertarget{normalized-ale-effect-size-measures}{%
\subsection{Normalized ALE effect size
measures}\label{normalized-ale-effect-size-measures}}

Since ALER and ALED scores are scaled on the range of \emph{y} for a
given dataset, these scores cannot be compared across datasets. Thus, we
present normalized versions of each with intuitive, comparable values.
For intuitive interpretation, we normalize the scores on the minimum,
median, and maximum of any dataset. In principle, we divide the
zero-centred \emph{y} values in a dataset into two halves: the lower
half from the 0th to the 50th percentile (the median) and the upper half
from the 50th to the 100th percentile. (Note that the median is included
in both halves). With zero-centred ALE \emph{y} values, all negative and
zero values are converted to their percentile score relative to the
lower half of the original \emph{y} values while all positive ALE
\emph{y} values are converted to their percentile score relative to the
upper half. (Technically, this percentile assignment is called the
empirical cumulative distribution function (ECDF) of each half.) Each
half is then divided by two to scale them from 0 to 50 so that together
they can represent 100 percentiles. (Note: when a centred ALE \emph{y}
value of exactly 0 occurs, we choose to include the score of zero ALE
\emph{y} in the lower half because it is analogous to the 50th
percentile of all values, which more intuitively belongs in the lower
half of 100 percentiles.) The transformed maximum ALE \emph{y} is then
scaled as a percentile from 0 to 100\%.

There is a notable complication. This normalization smoothly distributes
ALE \emph{y} values when there are many distinct values, but when there
are only a few distinct ALE \emph{y} values, then even a minimal ALE
\emph{y} deviation can have a relatively large percentile difference. If
any ALE \emph{y} value is less than the difference between the median in
the data and the value either immediately below or above the median, we
consider that it has virtually no effect. Thus, the normalization sets
such minimal ALE \emph{y} values as zero.

Its formula is in Equation~\ref{eq-norm_ale_y}.

\begin{equation}\protect\hypertarget{eq-norm_ale_y}{}{
norm\_ale\_y = 100 \times \begin{cases} 
0 & \text{if } \max(centred\_y < 0) \leq ale\_y \leq \min(centred\_y > 0), \\
\frac{-ECDF_{y_{\leq 0}}(ale\_y)}{2} & \text{if }ale\_y < 0 \\
\frac{ECDF_{y_{\geq 0}}(ale\_y)}{2} & \text{if }ale\_y > 0 \\
\end{cases} 
}\label{eq-norm_ale_y}\end{equation}

where - \(centred\_y\) is the vector of \texttt{y} values centred on the
median (that is, the median is subtracted from all values). -
\(ECDF_{y_{\geq 0}}\) is the ECDF of the non-negative values in
\texttt{y}. - \(-ECDF_{y_{\leq 0}}\) is the ECDF of the negative values
in \texttt{y} after they have been inverted (multiplied by -1).

Of course, the formula could be simplified by multiplying by 50 instead
of by 100 and not dividing the ECDFs by 2 each. But we prefer the form
we have given because it is explicit that each ECDF represents only half
the percentile range and that the result is scored to 100 percentiles.

\hypertarget{normalized-aler-naler}{%
\subsubsection{Normalized ALER (NALER)}\label{normalized-aler-naler}}

Based on this normalization, we first have the normalized ALER (NALER),
which scales the minimum and maximum ALE \emph{y} values from -50\% to
+50\%, centred on 0\%, which represents the median, as seen in
Equation~\ref{eq-naler}.

\begin{equation}\protect\hypertarget{eq-naler}{}{
\mathrm{NALER}(\mathrm{y, ale\_y}) = 
\{\min(\mathrm{norm\_ale\_y}) + 50, 
\max(\mathrm{norm\_ale\_y}) + 50 \}
}\label{eq-naler}\end{equation}

where \(y\) is the full vector of \emph{y} values in the original
dataset, required to calculate \(\mathrm{norm\_ale\_y}\).

ALER shows the extreme values of a variable's effect on the outcome. In
the effects plot above, it is indicated by the extreme ends of the
horizontal lines for each variable. We see there that \texttt{public}
has an ALER of -0.34, 0.42. When we consider that the median math score
in the dataset is 12.9, this ALER indicates that the minimum of any ALE
\emph{y} value for \texttt{public} (when \texttt{public\ ==\ TRUE}) is
-0.34 below the median. This is shown at the 12.6 mark in the plot
above. The maximum (\texttt{public\ ==\ FALSE}) is 0.42 above the
median, shown at the 13.3 point above. The ALER for \texttt{mean\_ses}
is considerably broader with -6.8 below and 2.6 above the median.

The result of this transformation is that NALER values can be
interpreted as percentile effects of \emph{y} below or above the median,
which is centred at 0\%. Their numbers represent the limits of the
effect of the \emph{x} variable with units in percentile scores of y. In
the effects plot above, because the percentile scale on the top
corresponds exactly to the raw scale below, the NALER limits are
represented by exactly the same points as the ALER limits; only the
scale changes. The scale for ALER and ALED is the lower scale of the raw
outcomes; the scale for NALER and NALED is the upper scale of
percentiles.

So, with a NALER of -4.26, 6.07, the minimum of any ALE value for
\texttt{public} (\texttt{public\ ==\ TRUE}) shifts math scores by -4
percentile \emph{y} points whereas the maximum
(\texttt{public\ ==\ FALSE}) shifts math scores by 6 percentile points.
The mean socioeconomic status has a NALER of -45.57, 32.66, ranging from
-46 to 33 percentile points of math scores.

\hypertarget{normalized-aled-naled}{%
\subsubsection{Normalized ALED (NALED)}\label{normalized-aled-naled}}

The normalization of ALED scores applies the same ALED formula as before
but on the normalized ALE values instead of on the original ALE \emph{y}
values, as seen in Equation~\ref{eq-naled}.

\begin{equation}\protect\hypertarget{eq-naled}{}{
\mathrm{NALED}(y, \mathrm{ale\_y}, \mathrm{ale\_n}) = \mathrm{ALED}(\mathrm{norm\_ale\_y}, \mathrm{ale\_n})
}\label{eq-naled}\end{equation}

NALED produces a score that ranges from 0 to 100\%. It is essentially
the ALED expressed in percentiles, that is, the average effect of a
variable over its full domain of values. So, the NALED of public school
status of 5 indicates that its average effect on math scores spans the
middle 5\% of scores. Mean socioeconomic status has an average effect
expressed in NALED of 12.7\% of scores.

The NALED is particularly helpful in comparing the practical relevance
of variables against our threshold by which we consider that a variable
needs to shift the outcome on average by more than 5\% of the median
values. This threshold is the same scale as the NALED. So, we can tell
that public school status with its NALED of 5 just barely crosses our
threshold.

\hypertarget{the-median-band-and-random-variables}{%
\subsection{The median band and random
variables}\label{the-median-band-and-random-variables}}

\begin{figure}

{\centering \includegraphics{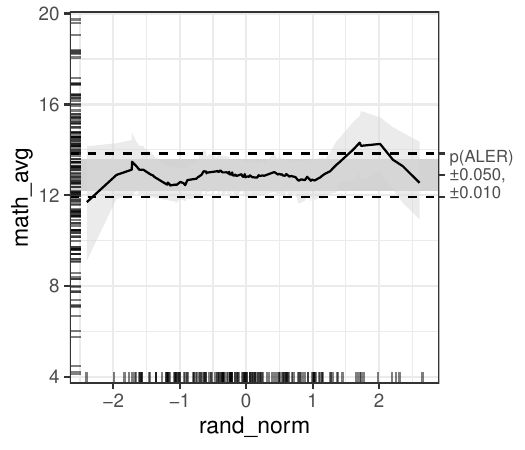}

}

\caption{\label{fig-math_random}ALE plot for the random variable from
mathematics achievement scores}

\end{figure}

It is particularly striking to focus on the random \texttt{rand\_norm}.
We reexamine its ALE plot in Figure~\ref{fig-math_random} and then we
examine its ALE effect size measures in
Table~\ref{tbl-math_ale_stats_rand_norm}.

\hypertarget{tbl-math_ale_stats_rand_norm}{}
\begin{longtable}[t]{>{}lrrrrrr}
\caption{\label{tbl-math_ale_stats_rand_norm}Effect size measures for the random variable from the math achievements
dataset }\tabularnewline

\toprule
statistic & estimate & p.value & conf.low & median & mean & conf.high\\
\midrule
\textbf{aled} & 0.358 & 0.000 & 0.108 & 0.361 & 0.358 & 0.582\\
\textbf{aler\_min} & -1.371 & 0.000 & -3.884 & -1.219 & -1.371 & -0.167\\
\textbf{aler\_max} & 1.546 & 0.000 & 0.315 & 1.566 & 1.546 & 2.764\\
\textbf{naled} & 4.774 & 0.010 & 1.519 & 4.688 & 4.774 & 8.454\\
\textbf{naler\_min} & -14.919 & 0.007 & -38.781 & -14.506 & -14.919 & -3.750\\
\addlinespace
\textbf{naler\_max} & 21.801 & 0.000 & 3.125 & 22.500 & 21.801 & 37.355\\
\bottomrule
\end{longtable}

\texttt{rand\_norm} has a NALED of 4.8. It might be surprising that a
purely random value has any ``effect size'' to speak of, but
statistically, it must have some numeric value or the other. In informal
tests with several different random seeds, the random variables never
exceeded this 5\% threshold. Thus, the effect of a variable like the
discrimination climate score (discrim, 4.7) should probably not be
considered practically meaningful.

\hypertarget{interpretation-of-normalized-ale-effect-sizes}{%
\subsection{Interpretation of normalized ALE effect
sizes}\label{interpretation-of-normalized-ale-effect-sizes}}

Here we summarize some general principles for interpreting normalized
ALE effect sizes.

\begin{itemize}
\tightlist
\item
  \textbf{Normalized ALE deviation (NALED):} this is the average
  variation of ALE effect of the input variable.

  \begin{itemize}
  \tightlist
  \item
    Values range from 0 to 100\%:

    \begin{itemize}
    \tightlist
    \item
      0\% means no effect at all.
    \item
      100\% means the maximum possible effect any variable could have:
      for a binary variable, one value (50\% of the data) sets the
      outcome at its minimum value and the other value (the other 50\%
      of the data) sets the outcome at its maximum value.
    \end{itemize}
  \item
    Larger NALED means stronger effects.
  \end{itemize}
\item
  \textbf{Normalized ALE range (NALER):} these are the minimum and
  maximum effects of any value of the input variable.

  \begin{itemize}
  \tightlist
  \item
    NALER minimum ranges from --50\% to 0\%; NALER maximum ranges from
    0\% to +50\%:
  \item
    0\% means no effect at all. It indicates that the only effect of the
    input variable is to keep the outcome at the median of its range of
    values.
  \item
    NALER minimum of \emph{n} means that, regardless of the effect size
    of NALED, the minimum effect of any input value shifts the outcome
    by \emph{n} percentile points of the outcome range. Lower values
    (closer to --50\%) mean a stronger extreme effect.
  \item
    NALER maximum of \emph{x} means that, regardless of the effect size
    of NALED, the maximum effect of any input value shifts the outcome
    by \emph{x} percentile points of the outcome range. Greater values
    (closer to +50\%) mean a stronger extreme effect.
  \end{itemize}
\end{itemize}

In general, regardless of the values of ALE statistics, we should always
visually inspect the ALE plots to identify and interpret patterns of
relationships between inputs and the outcome.

A common question for interpreting effect sizes is, ``How strong does an
effect need to be to be considered `strong' or `weak'?'' On one hand, we
refuse to offer general guidelines for how ``strong'' is ``strong''. The
simple answer is that it depends entirely on the applied context. It is
not meaningful to try to propose numerical values for statistics that
are supposed to be useful for all applied contexts.

On the other hand, we affirm the importance of delineating the threshold
between random effects and non-random effects. It is always important to
distinguish between a weak but real effect from one that is just a
statistical artifact due to random chance. For that, we propose that the
boundaries of ALER set by p-values should generally be used to determine
the acceptable risk of considering a statistic to be meaningful. First,
the analyst must select their acceptable p-values, understanding that
such a choice indicates their risk tolerance of confounding random
effects for true effects. So, if statistically significant ALE effects
are those that are less than the 0.05 p-value ALER minimum of a random
variable and greater than the 0.05 p-value maximum of a random variable,
this means that the analysis is willing to risk that 5\% of their
apparent effects might be partly do to random chance.

\hypertarget{statistical-inference-with-ale}{%
\section{Statistical inference with
ALE}\label{statistical-inference-with-ale}}

Although effect sizes are valuable in summarizing the global effects of
each variable, they mask much nuance, since each variable varies in its
effect along its domain of values. Thus, ALE is particularly powerful in
its ability to make fine-grained inferences of a variable's effect
depending on its specific value. Even more so than with effect sizes,
statistical inference is primarily a topic considered with relatively
smaller datasets. Thus, we demonstrate the principles in this section
only using the mathematics achievement dataset. That said, everything we
describe here is fully relevant to large datasets analyzed with ML.

\hypertarget{classical-statistical-inference}{%
\subsection{Classical statistical
inference}\label{classical-statistical-inference}}

\hypertarget{tbl-model_coefs}{}
\begin{longtable}[t]{>{}lrrrrrr}
\caption{\label{tbl-model_coefs}Bootstrapped coefficents for GAM for mathematics achievement scores }\tabularnewline

\toprule
term & estimate & conf.low & mean & median & conf.high & std.error\\
\midrule
\textbf{(Intercept)} & 12.713 & 11.654 & 12.713 & 12.718 & 13.613 & 0.484\\
\textbf{publicTRUE} & -0.689 & -2.015 & -0.689 & -0.652 & 0.415 & 0.637\\
\textbf{high\_minorityTRUE} & 1.031 & -0.318 & 1.031 & 1.054 & 2.324 & 0.676\\
\textbf{s(size)} & 3.536 & 1.000 & 3.536 & 2.503 & 8.385 & 2.603\\
\textbf{s(academic\_ratio)} & 6.085 & 1.000 & 6.085 & 7.323 & 8.756 & 2.810\\
\addlinespace
\textbf{s(female\_ratio)} & 4.203 & 1.000 & 4.203 & 3.652 & 8.396 & 2.382\\
\textbf{s(mean\_ses)} & 7.492 & 2.561 & 7.492 & 8.449 & 8.995 & 2.008\\
\textbf{s(minority\_ratio)} & 7.300 & 2.588 & 7.300 & 8.120 & 8.987 & 1.797\\
\textbf{s(discrim)} & 4.205 & 1.000 & 4.205 & 3.560 & 8.800 & 2.673\\
\textbf{s(rand\_norm)} & 6.796 & 1.000 & 6.796 & 7.777 & 8.767 & 2.394\\
\bottomrule
\end{longtable}

We begin by briefly reviewing the classical approach to statistical
inference. First, we can see the bootstrapped values of the effects of
individual variables in Table~\ref{tbl-model_coefs}. It is beyond the
scope of this article to explain in detail how GAM works (see Ross 2019
for a tutorial). However, for our model illustration here, the estimates
for the parametric variables (the non-numeric ones in our model) are
interpreted as regular statistical regression coefficients whereas the
estimates for the non-parametric smoothed variables (those whose
variable names are encapsulated by the smooth \texttt{s()} function) are
actually estimates for expected degrees of freedom (EDF in GAM). The
smooth function \texttt{s()} lets GAM model these numeric variables as
flexible curves that fit the data better than a straight line. The
\texttt{estimate} values for the smooth variables above are not so
straightforward to interpret, but suffice it to say that they are
completely different from regular regression coefficients.

With bootstrap-based confidence intervals, based on the default 95\%
confidence intervals, a coefficient is statistically significant if
\texttt{conf.low} and \texttt{conf.high} are both positive or both
negative. However, the statistical significance of the \texttt{estimate}
(EDF) of the smooth terms is meaningless here because EDF cannot go
below 1.0. Thus, even the random term \texttt{s(rand\_norm)} appears to
be ``statistically significant''. Only the values for the non-smooth
(parametric terms) \texttt{public} and \texttt{high\_minority} should be
considered here. So, we find that neither of the coefficient estimates
of \texttt{public} nor of \texttt{high\_minority} has an effect that is
statistically significantly different from zero. (The intercept is not
conceptually meaningful here; it is a statistical artifact.)

This initial analysis highlights two limitations of classical
hypothesis-testing analysis. First, it might work suitably well when we
use models that have traditional linear regression coefficients. But
once we use more advanced models like GAM that flexibly fit the data, we
cannot interpret coefficients meaningfully and so it is not so clear how
to reach inferential conclusions. Second, a basic challenge with models
that are based on the general linear model (including GAM and almost all
other statistical analyses) is that their coefficient significance
compares the estimates with the null hypothesis that there is no effect.
However, even if there is an effect, it might not be practically
meaningful. As we will see, ALE-based statistics are explicitly tailored
to emphasize practical implications beyond the notion of ``statistical
significance''.

\hypertarget{ale-data-structures-for-categorical-and-numeric-variables}{%
\subsection{ALE data structures for categorical and numeric
variables}\label{ale-data-structures-for-categorical-and-numeric-variables}}

To understand how bootstrapped ALE can be used for statistical
inference, we must understand the structure of ALE data. We can begin by
examining the structure for a binary variable with just two categories,
\texttt{public}, in Table~\ref{tbl-ale_data_public}.

\hypertarget{tbl-ale_data_public}{}
\begin{longtable}[t]{>{}lrrrr}
\caption{\label{tbl-ale_data_public}Structure of ALE data for \texttt{public} (categorical) from math
achievement dataset }\tabularnewline

\toprule
ale\_x & ale\_n & ale\_y & ale\_y\_lo & ale\_y\_hi\\
\midrule
\textbf{FALSE} & 70 & 13.302 & 12.192 & 14.176\\
\textbf{TRUE} & 90 & 12.613 & 11.892 & 13.350\\
\bottomrule
\end{longtable}

To understand how bootstrapped ALE can be used for statistical
inference, we must understand the structure of ALE data. We can begin by
examining the structure for a binary variable with just two categories,
\texttt{public}, in Table~\ref{tbl-ale_data_public}. The columns for a
categorical variable mean:

\begin{itemize}
\tightlist
\item
  \texttt{ale\_x}: the different categories that exist in the
  categorical variable.
\item
  \texttt{ale\_n}: the number of rows for that category in the dataset
  provided to the function.
\item
  \texttt{ale\_y}: the mean bootstrapped ALE function value calculated
  for that category.
\item
  \texttt{ale\_y\_lo} and \texttt{ale\_y\_hi}: the lower and upper
  confidence intervals for the bootstrapped \texttt{ale\_y} value.
\end{itemize}

By default, the \texttt{ale} package centres ALE values on the median of
the outcome variable; in our dataset, the median of all the schools'
average mathematics achievement scores is 12.9. With ALE centred on the
median, the weighted sum of ALE \emph{y} values (weighted on
\texttt{ale\_n}) above the median is approximately equal to the weighted
sum of those below the median. So, in the ALE plots above, when we
consider the number of instances indicated by the rug plots and category
percentages, the average weighted ALE \emph{y} approximately equals the
median.

\hypertarget{tbl-ale_data_mean_ses}{}
\begin{longtable}[t]{>{}rrrrr}
\caption{\label{tbl-ale_data_mean_ses}Structure of ALE data for \texttt{mean\_ses} (numeric) from math
achievement dataset }\tabularnewline

\toprule
ale\_x & ale\_n & ale\_y & ale\_y\_lo & ale\_y\_hi\\
\midrule
\textbf{-1.188} & 1 & 5.714 & 2.662 & 9.535\\
\textbf{-1.043} & 1 & 7.574 & 5.753 & 10.194\\
\textbf{-0.792} & 2 & 10.045 & 8.291 & 11.970\\
\textbf{-0.756} & 1 & 10.261 & 8.439 & 11.813\\
\textbf{-0.703} & 2 & 10.485 & 8.672 & 11.852\\
\addlinespace
\textbf{-0.699} & 1 & 10.618 & 9.214 & 11.966\\
\textbf{-0.674} & 2 & 10.806 & 9.476 & 12.158\\
\textbf{-0.663} & 2 & 10.883 & 9.560 & 12.277\\
\textbf{-0.643} & 1 & 10.670 & 9.549 & 11.758\\
\textbf{-0.591} & 2 & 11.172 & 10.284 & 12.243\\
\addlinespace
\textbf{...} & ... & ... & ... & ...\\
\textbf{0.535} & 2 & 14.022 & 13.107 & 14.932\\
\textbf{0.569} & 2 & 14.342 & 13.309 & 15.205\\
\textbf{0.617} & 1 & 14.617 & 13.696 & 15.768\\
\textbf{0.633} & 2 & 14.731 & 13.477 & 15.616\\
\addlinespace
\textbf{0.657} & 1 & 15.097 & 13.786 & 16.482\\
\textbf{0.666} & 2 & 15.039 & 13.769 & 16.552\\
\textbf{0.688} & 2 & 15.158 & 13.708 & 16.747\\
\textbf{0.718} & 1 & 15.277 & 13.746 & 17.002\\
\textbf{0.759} & 2 & 15.420 & 13.572 & 18.007\\
\addlinespace
\textbf{0.831} & 1 & 14.611 & 12.073 & 16.320\\
\bottomrule
\end{longtable}

In Table~\ref{tbl-ale_data_mean_ses}, we see the ALE data structure for
a numeric variable, \texttt{mean\_ses}. The columns are the same as with
a categorical variable, but the meaning of \texttt{ale\_x} is different
since there are no categories. To calculate ALE for numeric variables,
the range of \emph{x} values is divided into fixed intervals (such as
into 100 percentiles). If the \emph{x} values have fewer than 100
distinct values in the data, then each distinct value becomes an ale\_x
interval. (This is often the case with smaller datasets like ours; here
\texttt{mean\_ses} has only 97 distinct values.) If there are more than
100 distinct values, then the range is divided into 100 percentile
groups. So, \texttt{ale\_x} represents each of these x-variable
intervals. The other columns mean the same thing as with categorical
variables: \texttt{ale\_n} is the number of rows of data in each
\texttt{ale\_x} interval and \texttt{ale\_y} is the calculated ALE for
that \texttt{ale\_x} value.

\hypertarget{inference-with-ale-based-on-bootstrapped-confidence-intervals}{%
\subsection{Inference with ALE based on bootstrapped confidence
intervals}\label{inference-with-ale-based-on-bootstrapped-confidence-intervals}}

Whereas the coefficient table above based on classic statistics
indicated this conclusion for \texttt{public}, it indicated that
\texttt{high\_minority} had a statistically significant effect; our ALE
analysis indicates that \texttt{high\_minority} does not. In addition, e

With the structure of ALE data clear, we can now proceed to statistical
inference with ALE based on bootstrapped confidence intervals. In a
bootstrapped ALE plot, values within the confidence intervals are
statistically significant; values outside of the ALER band can be
considered at least somewhat meaningful. Thus, \textbf{the essence of
ALE-based statistical inference is that only effects that are
simultaneously within the confidence intervals AND outside of the ALER
band should be considered conceptually meaningful.}

\begin{figure}

{\centering \includegraphics{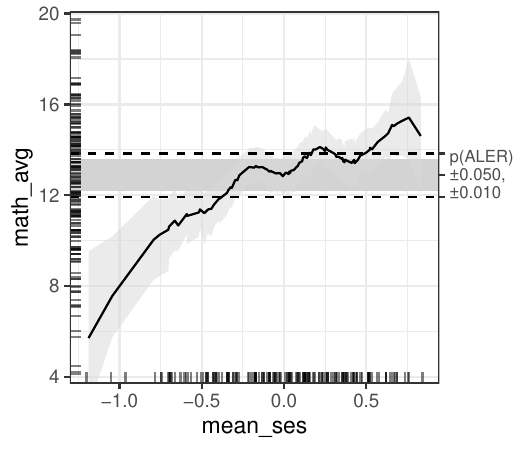}

}

\caption{\label{fig-math_mean_ses}ALE plot for \texttt{mean\_ses} from
math achievement dataset}

\end{figure}

\hypertarget{tbl-conf_mean_ses}{}
\begin{longtable}[t]{>{}r>{}rrrrrrr>{}l}
\caption{\label{tbl-conf_mean_ses}Confidence regions for \texttt{mean\_ses} (numeric) from math
achievement dataset }\tabularnewline

\toprule
start\_x & end\_x & x\_span & n & n\_pct & start\_y & end\_y & trend & relative\_to\_mid\\
\midrule
\textbf{-1.188} & \textbf{-0.674} & 0.255 & 10 & 0.062 & 5.714 & 10.806 & 1.343 & \textbf{below}\\
\textbf{-0.663} & \textbf{-0.663} & 0.000 & 2 & 0.013 & 10.883 & 10.883 & 0.000 & \textbf{overlap}\\
\textbf{-0.643} & \textbf{-0.643} & 0.000 & 1 & 0.006 & 10.670 & 10.670 & 0.000 & \textbf{below}\\
\textbf{-0.591} & \textbf{-0.591} & 0.000 & 2 & 0.013 & 11.172 & 11.172 & 0.000 & \textbf{overlap}\\
\textbf{-0.588} & \textbf{-0.588} & 0.000 & 1 & 0.006 & 11.093 & 11.093 & 0.000 & \textbf{below}\\
\addlinespace
\textbf{-0.517} & \textbf{-0.511} & 0.003 & 4 & 0.025 & 11.247 & 11.377 & 2.930 & \textbf{overlap}\\
\textbf{-0.484} & \textbf{-0.484} & 0.000 & 1 & 0.006 & 11.212 & 11.212 & 0.000 & \textbf{below}\\
\textbf{-0.467} & \textbf{0.569} & 0.513 & 127 & 0.794 & 11.337 & 14.342 & 0.393 & \textbf{overlap}\\
\textbf{0.617} & \textbf{0.617} & 0.000 & 1 & 0.006 & 14.617 & 14.617 & 0.000 & \textbf{above}\\
\textbf{0.633} & \textbf{0.633} & 0.000 & 2 & 0.013 & 14.731 & 14.731 & 0.000 & \textbf{overlap}\\
\addlinespace
\textbf{0.657} & \textbf{0.718} & 0.030 & 6 & 0.038 & 15.097 & 15.277 & 0.400 & \textbf{above}\\
\textbf{0.759} & \textbf{0.831} & 0.036 & 3 & 0.019 & 15.420 & 14.611 & -1.523 & \textbf{overlap}\\
\bottomrule
\end{longtable}

We can see this, for example, with the plot of \texttt{mean\_ses} in
Figure~\ref{fig-math_mean_ses}. However, it might not always be easy to
tell from a plot which regions are relevant, so the results of
statistical significance are summarized with a confidence regions table
in Table~\ref{tbl-conf_mean_ses}. For numeric variables, the confidence
regions summary has one row for each consecutive sequence of \emph{x}
values that have the same status: all values in the region are below the
ALER band, they overlap the band, or they are all above the band. Here
are the summary components:

\begin{itemize}
\tightlist
\item
  \texttt{start\_x} is the first and \texttt{end\_x} is the last
  \emph{x} value in the sequence. \texttt{start\_y} is the \emph{y}
  value that corresponds to \texttt{start\_x} while \texttt{end\_y}
  corresponds to \texttt{end\_x}.
\item
  \texttt{n} is the number of data elements in the sequence;
  \texttt{n\_pct} is the percentage of total data elements out of the
  total number.
\item
  \texttt{x\_span} is the length of \emph{x} of the sequence that has
  the same confidence status. However, so that it may be comparable
  across variables with different units of x, \texttt{x\_span} is
  expressed as a percentage of the full domain of \emph{x} values.
\item
  \texttt{trend} is the average slope from the point
  \texttt{(start\_x,\ start\_y)} to \texttt{(end\_x,\ end\_y)}. Because
  only the start and end points are used to calculate \texttt{trend}, it
  does not reflect any ups and downs that might occur between those two
  points. Since the various \emph{x} values in a dataset are on
  different scales, the scales of the \emph{x} and \emph{y} values in
  calculating the \texttt{trend} are normalized on a scale of 0 to 1
  each so that the trends for all variables are directly comparable. A
  positive \texttt{trend} means that, on average, \emph{y} increases
  with \emph{x}; a negative \texttt{trend} means that, on average,
  \emph{y} decreases with x; a zero \texttt{trend} means that \emph{y}
  has the same value at its start and end points--this is always the
  case if there is only one point in the indicated sequence.
\item
  \textbf{\texttt{relative\_to\_mid}} is the key information here. It
  indicates if all the values in sequence from \texttt{start\_x} to
  \texttt{end\_x} are below, overlapping, or above the ALER band:

  \begin{itemize}
  \tightlist
  \item
    \textbf{below}: the higher limit of the confidence interval of ALE
    \emph{y} (\texttt{ale\_y\_hi}) is below the lower limit of the ALER
    band.
  \item
    \textbf{above}: the lower limit of the confidence interval of ALE
    \emph{y} (\texttt{ale\_y\_lo}) is above the higher limit of the ALER
    band.
  \item
    \textbf{overlap}: neither of the first two conditions holds; that
    is, the confidence region from \texttt{ale\_y\_lo} to
    \texttt{ale\_y\_hi} at least partially overlaps the ALER band.
  \end{itemize}
\end{itemize}

These results tell us simply that, for \texttt{mean\_ses}, from -1.19 to
-1.04, ALE is below the median band from 6.1 to 7.6. From -0.792 to
-0.792, ALE overlaps the median band from 10.2 to 10.2. From -0.756 to
-0.674, ALE is below the median band from 10.2 to 10.8. From -0.663 to
-0.663, ALE overlaps the median band from 10.9 to 10.9. From -0.643 to
-0.484, ALE is below the median band from 10.8 to 11.2. From -0.467 to
-0.467, ALE overlaps the median band from 11.5 to 11.5. From -0.46 to
-0.46, ALE is below the median band from 11.4 to 11.4. A few other
regions briefly exceeded the ALER band.

Considering the details from Table~\ref{tbl-conf_mean_ses}, we can see
that socioeconomic status overlaps with the median range for most of its
range. However, although there are several ups and downs with the
bootstrapped ranges, most of the schools with \texttt{mean\_ses} scores
below -0.46

\begin{figure}

{\centering \includegraphics{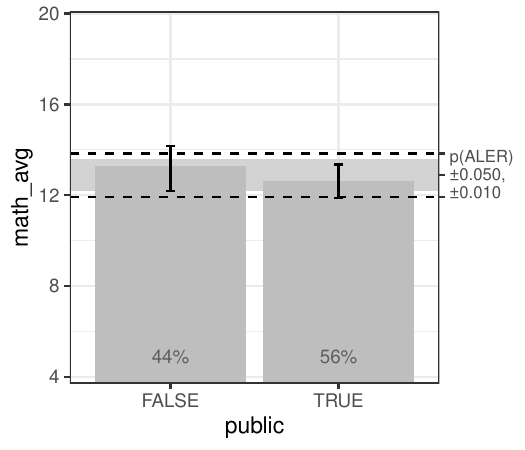}

}

\caption{\label{fig-math_public}ALE plot for \texttt{public} from math
achievement dataset}

\end{figure}

\hypertarget{tbl-conf_public}{}
\begin{longtable}[t]{>{}lrrr>{}l}
\caption{\label{tbl-conf_public}Confidence regions for \texttt{public} (categorical) from math
achievement dataset }\tabularnewline

\toprule
x & n & n\_pct & y & relative\_to\_mid\\
\midrule
\textbf{FALSE} & 70 & 0.438 & 13.302 & \textbf{overlap}\\
\textbf{TRUE} & 90 & 0.562 & 12.613 & \textbf{overlap}\\
\bottomrule
\end{longtable}

Confidence region summary tables are available not only for numeric but
also for categorical variables, as we see with the ALE plot for
\texttt{public} (Figure~\ref{fig-math_public}) and its confidence
regions summary table in Table~\ref{tbl-conf_public}. Since we have
categories here, there are no start or end positions and there is no
trend. We instead have each \texttt{x} category and its single ALE
\texttt{y} value, with the \texttt{n} and \texttt{n\_pct} of the
respective category and \texttt{relative\_to\_mid} as before to indicate
whether the indicated category is below, overlaps with, or is above the
ALER band. These results tell us that, for \texttt{public}, for FALSE,
the ALE of 13.3 overlaps the ALER band. For TRUE, the ALE of 12.6 also
overlaps the ALER band. In other words, there is no statistically (or
practically) significant value of \texttt{public}. The data does not
provide evidence to support the claim that public or non-public schools
have significantly different math achievement scores.

\hypertarget{confidence-regions-and-random-variables}{%
\subsection{Confidence regions and random
variables}\label{confidence-regions-and-random-variables}}

\begin{figure}

{\centering \includegraphics{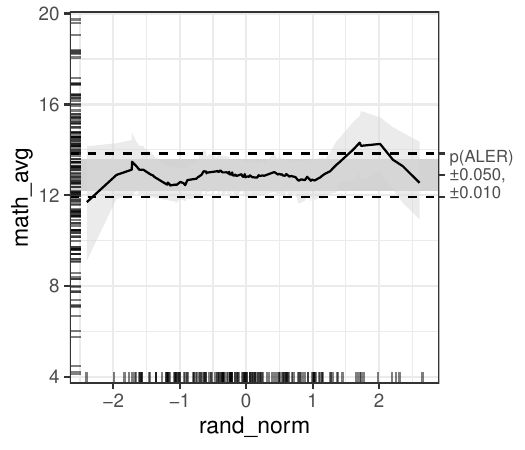}

}

\caption{\label{fig-math_rand_norm}ALE plot for the random variable from
math achievement dataset}

\end{figure}

\hypertarget{tbl-math_conf_rand_norm}{}
\begin{longtable}[t]{>{}r>{}rrrrrrr>{}l}
\caption{\label{tbl-math_conf_rand_norm}Confidence regions for the full-model (appropriately) bootstrapped
random variable from math achievement dataset }\tabularnewline

\toprule
start\_x & end\_x & x\_span & n & n\_pct & start\_y & end\_y & trend & relative\_to\_mid\\
\midrule
\textbf{-2.397} & \textbf{2.608} & 1 & 160 & 1 & 11.699 & 12.55 & 0.057 & \textbf{overlap}\\
\bottomrule
\end{longtable}

Again, our random variable \texttt{rand\_norm} is particularly
interesting, as we can see from its ALE plot in
Figure~\ref{fig-math_rand_norm} and its confidence regions summary table
in Table~\ref{tbl-math_conf_rand_norm}. Despite the apparent pattern, we
see that from -2.4 to 2.61, ALE overlaps the ALER band from 11.7 to
12.6. So, despite the random highs and lows in the bootstrap confidence
interval, there is no reason to suppose that the random variable has any
effect anywhere in its domain.

\hypertarget{discussion}{%
\section{Discussion}\label{discussion}}

In the preceding sections, we have navigated through the multifaceted
landscape of ALE, exploring its potential and the innovative extensions
proposed in this study. The contributions we present not only forge a
path towards more nuanced, interpretable, and computationally efficient
applications in ML model interpretation. Through the implementation of
full model bootstrapping, the introduction of intuitive ALE-based effect
size measures, and the innovative concept of confidence regions, this
work endeavours to enhance the robustness and interpretability of ALE
analyses.

In this concluding discussion, we shall reflect on the implications,
applications, and potential future trajectories of these contributions,
anchoring our discourse in the context of both the theoretical and
practical domains of ML model interpretation and analysis.

\hypertarget{contributions}{%
\subsection{Contributions}\label{contributions}}

The extensions to the implementation of ALE that we have described in
the preceding sections address the opportunities for improvement that we
highlight above in the section on related work.

Whereas data-only bootstrapping of ALE values has already been
implemented

First, we implement full model bootstrapping specifically tailored for
calculating ALE rather than just data-only bootstrapping. This allows
the creation of appropriate confidence intervals for the ALE of models
that are too small to be fine-tuned with ML data validation techniques.
It is not surprising that the Python packages that implement
bootstrapping or variations feature just data-only bootstrapping (Flora
2023; Jumelle, Kuhn-Regnier and Rajaratnam 2020). Python is rarely used
to analyze small datasets that are not amenable to a train-test split;
smaller datasets are typically analyzed with R rather than with Python.

Second, we introduce two new pairs of effect size measures based on ALE
designed for intuitive interpretability. Unlike the overall model MEC
and the IAS (Molnar, Casalicchio and Bischl 2020) that only show the
average effects across all variables, the measures we present here show
the effects for individual variables. The ALE deviation indicates the
average dispersion of ALE values while the sALE range indicates the
maximum dispersion. These base measures are scaled to the \emph{y}
outcome variable and so are readily interpretable in terms of the
applications to the outcome; they are also directly comparable across
predictor variables. For meaningful comparison of effect sizes across
different datasets and contexts, each of these has a normalized version
that is scaled on a percentile scale relative to the range of possible
\emph{y} outcome values.

Third, to address the nuances of non-linear relationships, we go beyond
effect size measures and introduce the notion of confidence regions as
an approach to statistical inference based on ALE confidence intervals.
Whereas several ML effect size measures like the impact score (Lötsch
and Ultsch 2020) and \(f^2_v\) (Messner 2023)---and, indeed, ALED, ALER,
NALED, and NALER---only indicate a general effect, Messner (2023) has
made an initial effort in indicating the complexity of relationships by
testing for monotonicity and average direction. Confidence regions fully
embrace the potential complexity of relationships by not trying to
reduce them to a single number. Rather, they identify areas
significantly above or below a ALER band. For numeric predictors,
confidence regions indicate the trend within these regions to infer the
average slope, somewhat mirroring Messner's slope measure, but on a more
granular scale that communicates necessary nuances. Monotonicity is less
pertinent from this perspective; understanding the relationship's
intricate shape, which necessitates graphical interpretation, holds
greater significance than a single numeric summary.

A fourth minor but non-negligible contribution is that, as these
measures and techniques are all calculated directly from the data
generated from ALE, they not only inherit the inherent superior
computational efficiency compared to other processor-intensive effect
size measures (Molnar 2022), but they are computationally negligible to
calculate in any analysis that has already calculated ALE.

\hypertarget{practical-implications}{%
\subsection{Practical Implications}\label{practical-implications}}

There are several practical implications of our ALE-based effect size
measures and of statistical inference using bootstrapped ALE confidence
regions.

Overall effect size measures afford a comprehensive perspective on model
performance, providing a global depiction of the impact of each variable
on the outcome throughout the model. In contrast, due to the
non-homogeneous effect across the entire domain of the \emph{x} input
factors, the ALE confidence regions delineate specific zones within the
domain of \emph{x} where an input predictor may be active or inactive.

The confidence regions offer a nuanced context to aid the interpretation
of the overall measures. Specifically, when encountering a particularly
broad ALE range or a substantial ALE range interval, coupled with a more
modest ALE deviation score, the confidence regions clarify which areas
of the domain exhibit atypical effects of that variable.

Overall effect size measures facilitate rapid identification of
variables that most significantly impact the outcomes of a system of
interest. Subsequently, for precise action, the confidence regions
assist decision-makers in focusing on specific values or ranges of
interest that might exert undesirable or desirable effects on the
outcome.

The intriguing occurrence of a substantial disparity between overall
effects, as indicated by ALE deviation, and extreme effects, as
highlighted by ALE range, can be dissected with the assistance of
confidence regions. These focus on the origins of such disparities and,
instead of being deemed problematic, can serve as a fountainhead of
unexpected insights. Given that ALE averages effects across intervals,
spikes in the ALE range do not invariably represent the influence of
potential spurious outliers; they typically signify the effects within a
particular data range. Such disparities generally merit closer
investigation, often paving the way for unforeseen and valuable
insights.

Generally, a pitfall of any overall effect size measure is that it may
misleadingly convey the data as more homogeneous than it truly is. While
some extensions have been proposed to the original ALE algorithm to
account for such data heterogeneity, the confidence regions adeptly
illuminate them by indicating where effects are homogeneous and where
they are heterogeneous.

\hypertarget{conclusion}{%
\subsection{Conclusion}\label{conclusion}}

In this article, we have unveiled novel methodologies and metrics that
enhance the applicability and interpretability of ALE in ML model
analysis. Our contributions span the implementation of a full model
bootstrapping approach, tailored for ALE, which facilitates the creation
of apt confidence intervals, especially pivotal for models trained on
smaller datasets. Furthermore, we introduced two pairs of effect size
measures---ALE deviation (ALED) and ALE range (ALER), along with their
normalized counterparts (NALED and NALER)---designed to provide a
nuanced depiction of individual variable effects, thereby offering a
more granular insight compared to existing models. The introduction of
confidence regions, which embrace the complexity of relationships
without reducing them to a singular numeric value, marks a significant
stride towards comprehending the intricacies of non-linear
relationships. Lastly, the computational efficiency of our proposed
measures and techniques, derived directly from ALE data, not only
underscores their practicality but also positions them as
computationally viable options in analyses that have already computed
ALE.

As we reflect upon these contributions, we recognize the potential they
hold in bridging theoretical understanding and practical application,
thereby paving the way for more robust, interpretable, and insightful
analyses in the realm of ML model interpretation. Future endeavours may
explore further applications and validations of these methodologies
across diverse datasets and contexts, ensuring their robustness and
utility in varied analytical scenarios.

\hypertarget{acknowledgments}{%
\section{Acknowledgments}\label{acknowledgments}}

We used ChatGPT in developing the outline of this article and in
drafting much of the text. The author takes full responsibility for the
entirety of the final revised contents. In addition, we used ChatGPT
extensively in developing the code for the \texttt{ale} package.

\hypertarget{references}{%
\section{References}\label{references}}

\hypertarget{refs}{}
\begin{CSLReferences}{1}{0}
\leavevmode\vadjust pre{\hypertarget{ref-apley2018}{}}%
Apley, Daniel W. 2018. \emph{{ALEPlot}: {Accumulated Local Effects}
({ALE}) {Plots} and {Partial Dependence} ({PD}) {Plots}}. 2018. 1.1.
\textless{}\url{https://cran.r-project.org/web/packages/ALEPlot/}\textgreater{}
Retrieved June 28, 2023.

\leavevmode\vadjust pre{\hypertarget{ref-apley2020}{}}%
Apley, Daniel W. and Jingyu Zhu. 2020. Visualizing the {Effects} of
{Predictor Variables} in {Black Box Supervised Learning Models}.
\emph{Journal of the Royal Statistical Society Series B: Statistical
Methodology} 82, 4, 1059--1086.
\textless{}\url{https://doi.org/10.1111/rssb.12377}\textgreater{}
Retrieved June 16, 2023.

\leavevmode\vadjust pre{\hypertarget{ref-biecek2023}{}}%
Biecek, Przemyslaw, Szymon Maksymiuk and Hubert Baniecki. 2023.
\emph{{DALEX}: {moDel Agnostic Language} for {Exploration} and
{eXplanation}}. 2023. 2.4.3.
\textless{}\url{https://cran.r-project.org/web/packages/DALEX/index.html}\textgreater{}
Retrieved October 10, 2023.

\leavevmode\vadjust pre{\hypertarget{ref-flora2023}{}}%
Flora, Montgomery. 2023. \emph{Scikit-explain: {A} user-friendly python
package for computing and plotting machine learning explainability
output.} 2023. 0.1.4.
\textless{}\url{https://github.com/monte-flora/scikit-explain/}\textgreater{}
Retrieved October 10, 2023.

\leavevmode\vadjust pre{\hypertarget{ref-friedman2008}{}}%
Friedman, Jerome H. and Bogdan E. Popescu. 2008. Predictive {Learning}
via {Rule Ensembles}. \emph{The Annals of Applied Statistics} 2, 3,
916--954.
\textless{}\url{https://www.jstor.org/stable/30245114}\textgreater{}
Retrieved October 12, 2023.

\leavevmode\vadjust pre{\hypertarget{ref-gkolemis2023a}{}}%
Gkolemis, Vasilis et al. 2023. {RHALE}: {Robust} and
{Heterogeneity-aware Accumulated Local Effects}.
\textless{}\url{http://arxiv.org/abs/2309.11193}\textgreater{} Retrieved
October 5, 2023.

\leavevmode\vadjust pre{\hypertarget{ref-gkolemis2023}{}}%
Gkolemis, Vasilis, Theodore Dalamagas and Christos Diou. 2023. {DALE}:
{Differential Accumulated Local Effects} for efficient and accurate
global explanations. In \emph{Proceedings of {The} 14th {Asian
Conference} on {Machine} {Learning}}, 375--390. {PMLR}.
\textless{}\url{https://proceedings.mlr.press/v189/gkolemis23a.html}\textgreater{}
Retrieved October 5, 2023.

\leavevmode\vadjust pre{\hypertarget{ref-jomar2023}{}}%
Jomar, Dana. 2023. \emph{{PyALE}: {ALE} plots with python}. 2023. 1.1.3.
\textless{}\url{https://github.com/DanaJomar/PyALE}\textgreater{}
Retrieved October 10, 2023.

\leavevmode\vadjust pre{\hypertarget{ref-jumelle2020}{}}%
Jumelle, Maxime, Alexander Kuhn-Regnier and Sanjif Rajaratnam. 2020.
\emph{{ALEPython}}. 2020.
\textless{}\url{https://github.com/blent-ai/ALEPython}\textgreater{}
Retrieved October 10, 2023.

\leavevmode\vadjust pre{\hypertarget{ref-lotsch2020}{}}%
Lötsch, Jörn and Alfred Ultsch. 2020. A non-parametric effect-size
measure capturing changes in central tendency and data distribution
shape. \emph{PLOS ONE} 15, 9, e0239623.
\textless{}\url{https://journals.plos.org/plosone/article?id=10.1371/journal.pone.0239623}\textgreater{}
Retrieved October 12, 2023.

\leavevmode\vadjust pre{\hypertarget{ref-mayer2021}{}}%
Mayer, Michael. 2021. A {Curious Fact} on the {Diamonds Dataset} --
{Michael}'s and {Christian}'s {Blog}.
\textless{}\url{https://lorentzen.ch/index.php/2021/04/16/a-curious-fact-on-the-diamonds-dataset/}\textgreater{}
Retrieved October 3, 2023.

\leavevmode\vadjust pre{\hypertarget{ref-messner2023}{}}%
Messner, Wolfgang. 2023. From black box to clear box: {A} hypothesis
testing framework for scalar regression problems using deep artificial
neural networks. \emph{Applied Soft Computing} 146, 110729.
\textless{}\url{https://www.sciencedirect.com/science/article/pii/S1568494623007470}\textgreater{}
Retrieved October 11, 2023.

\leavevmode\vadjust pre{\hypertarget{ref-molnar2022}{}}%
Molnar, Christoph. 2022. \emph{Interpretable {Machine Learning}}.
\textless{}\url{https://christophm.github.io/interpretable-ml-book/}\textgreater{}
Retrieved January 17, 2019.

\leavevmode\vadjust pre{\hypertarget{ref-molnar2020}{}}%
Molnar, Christoph, Giuseppe Casalicchio and Bernd Bischl. 2020.
\href{https://doi.org/10.1007/978-3-030-43823-4_17}{Quantifying {Model
Complexity} via {Functional Decomposition} for {Better Post-hoc
Interpretability}}. In P. Cellier and K. Driessens (eds.), \emph{Machine
{Learning} and {Knowledge Discovery} in {Databases}}, Communications in
{Computer} and {Information Science}, 193--204. {Cham}: {Springer
International Publishing}.

\leavevmode\vadjust pre{\hypertarget{ref-molnar2022a}{}}%
Molnar, Christoph and Patrick Schratz. 2022. \emph{Iml: {Interpretable
Machine Learning}}. 2022. 0.11.1.
\textless{}\url{https://cran.r-project.org/web/packages/iml/}\textgreater{}
Retrieved October 10, 2023.

\leavevmode\vadjust pre{\hypertarget{ref-okoli2023a}{}}%
Okoli, Chitu. 2023. \emph{Ale: {Interpretable Machine Learning} and
{Statistical Inference} with {Accumulated Local Effects} ({ALE})}. 2023.
0.2.0.
\textless{}\url{https://cran.r-project.org/web/packages/ale/index.html}\textgreater{}
Retrieved October 10, 2023.

\leavevmode\vadjust pre{\hypertarget{ref-pinheiro2023}{}}%
Pinheiro, José et al. 2023. \emph{Nlme: {Linear} and {Nonlinear Mixed
Effects Models}}. 2023. 3.1-163.
\textless{}\url{https://cran.r-project.org/web/packages/nlme/}\textgreater{}
Retrieved October 13, 2023.

\leavevmode\vadjust pre{\hypertarget{ref-rapidminer2023}{}}%
RapidMiner. 2023. \emph{Interpretation}. 2023. 0.7.1.
\textless{}\url{https://marketplace.rapidminer.com/UpdateServer/faces/product_details.xhtml?productId=rmx_interpretation}\textgreater{}
Retrieved October 10, 2023.

\leavevmode\vadjust pre{\hypertarget{ref-ross2019}{}}%
Ross, Noam. 2019. Introduction to {Generalized Additive Models} ·
{Generalized Additive Models} in {R}.
\textless{}\url{https://noamross.github.io/gams-in-r-course/chapter1}\textgreater{}
Retrieved October 13, 2023.

\leavevmode\vadjust pre{\hypertarget{ref-sechidis2015}{}}%
Sechidis, Konstantinos. 2015. \emph{Hypothesis testing and feature
selection in semi-supervised data}. PhD thesis. {University of
Manchester}.
\textless{}\url{https://www.research.manchester.ac.uk/portal/en/theses/hypothesis-testing-and-feature-selection-in-semisupervised-data(97f5f950-f020-4ace-b6cd-49cb2f88c730).html}\textgreater{}
Retrieved October 12, 2023.

\leavevmode\vadjust pre{\hypertarget{ref-seldontechnologies2023}{}}%
Seldon Technologies. 2023. \emph{Alibi: {Algorithms} for monitoring and
explaining machine learning models}. 2023. 0.9.4.
\textless{}\url{https://github.com/SeldonIO/alibi}\textgreater{}
Retrieved October 10, 2023.

\leavevmode\vadjust pre{\hypertarget{ref-tong2017}{}}%
Tong, L. I., R. Saminathan and C. W. Chang. 2017. Uncertainty
{Assessment} of {Non-normal Emission Estimates Using Non-parametric
Bootstrap Confidence Intervals}. \emph{Journal of Environmental
Informatics} 28, 1, 1, 61--70.
\textless{}\url{http://www.jeionline.org/index.php?journal=mys\&page=article\&op=view\&path\%5B\%5D=201500322}\textgreater{}
Retrieved October 11, 2023.

\leavevmode\vadjust pre{\hypertarget{ref-wickham2023}{}}%
Wickham, Hadley et al. 2023. \emph{Ggplot2: {Create Elegant Data
Visualisations Using} the {Grammar} of {Graphics}}. 2023. 3.4.4.
\textless{}\url{https://cran.r-project.org/web/packages/ggplot2/index.html}\textgreater{}
Retrieved October 13, 2023.

\leavevmode\vadjust pre{\hypertarget{ref-wood2023}{}}%
Wood, Simon. 2023. \emph{Mgcv: {Mixed GAM Computation Vehicle} with
{Automatic Smoothness Estimation}}. 2023. 1.9-0.
\textless{}\url{https://cran.r-project.org/web/packages/mgcv/}\textgreater{}
Retrieved October 13, 2023.

\end{CSLReferences}

\end{document}